%% file: vamoh.tex
\documentclass{article}

\usepackage{microtype}
\usepackage[final,pdftex]{graphicx}      %
\usepackage{caption}
\usepackage{subcaption}
\usepackage{booktabs} %
\usepackage{amsmath,amsfonts,amsthm,bm}       %
\usepackage{stmaryrd} %

\usepackage{hyperref}

\usepackage{diagbox}
\usepackage[accepted]{icml2023}

\usepackage{amsmath}
\usepackage{amssymb}
\usepackage{mathtools}
\usepackage{amsthm}

\usepackage[capitalize,noabbrev]{cleveref}

\usepackage{preamble}  %
\theoremstyle{plain}

\theoremstyle{definition}

\theoremstyle{remark}

\usepackage{enumitem}

\icmltitlerunning{Variational Mixture of HyperGenerators for Learning Distributions over Functions}

\begin{document}

\twocolumn[
\icmltitle{Variational Mixture of HyperGenerators \\for Learning Distributions over Functions}

\icmlsetsymbol{equal}{*}

\begin{icmlauthorlist}
\icmlauthor{Batuhan Koyuncu}{saar}
\icmlauthor{Pablo S\'anchez-Mart\'in}{imprs}
\icmlauthor{Ignacio Peis}{madrid}
\icmlauthor{Pablo M. Olmos}{madrid}
\icmlauthor{Isabel Valera}{saar}
\end{icmlauthorlist}

\icmlaffiliation{saar}{Saarland University, Saarbrücken, Germany}
\icmlaffiliation{imprs}{Max Planck Institute for Intelligent Systems, Tübingen, Germany}
\icmlaffiliation{madrid}{Universidad Carlos III de Madrid, Madrid, Spain}

\icmlcorrespondingauthor{Batuhan Koyuncu}{koyuncu@cs.uni-saarland.de}
\icmlkeywords{Machine Learning, ICML}

\vskip 0.3in
]

\printAffiliationsAndNotice{}  %

\begin{abstract}
Recent approaches build on 
implicit neural representations (INRs) to propose generative models over function spaces. 
However, they are computationally costly when dealing with inference tasks, 
such as missing data imputation, or directly cannot tackle them. 
In this work, we propose a novel deep generative model, named \ours. \ours\ combines the capabilities of modeling continuous functions using INRs and the inference capabilities of Variational Autoencoders (VAEs).
In addition, \ours\ relies on a normalizing flow to define the prior, and a mixture of hypernetworks to parametrize the data log-likelihood. This gives \ours\ a high expressive capability and interpretability.
Through experiments on a diverse range of data types, such as images, voxels, and climate data, we show that \ours\ can effectively learn rich distributions over continuous functions. Furthermore, it can perform inference-related tasks, such as conditional super-resolution generation and in-painting, as well or better than previous approaches, while being less computationally demanding.
\end{abstract}

\input{sections/intro.tex}
\input{sections/related_work.tex}
\input{sections/our_model_v3.tex}

\input{sections/training.tex}
\input{sections/experiments.tex}

\input{sections/conclusions.tex}
\input{sections/acknow}
\newpage

\bibliography{vamoh}
\bibliographystyle{icml2023}

\newpage
\appendix
\onecolumn

\input{appendices/our_model_extended.tex}
\input{appendices/experiments_extended.tex}

\end{document}

%% file: sections/intro.tex
\section{Introduction}

While many real-world applications lead to data over continuous coordinate systems, such data is often discretized, e.g., by fixing the resolution of images \cite{simonyan2014very} or assuming a fixed sample frequency in time-series \cite{hochreiter1997long}. 
In contrast, recent advances in Implicit Neural Representations (INR) have been shown to  be powerful approaches for directly parameterizing  continuous functions by mapping coordinates into data features. 
To name a few examples, INRs have been successfully applied in diverse fields such as image representation \cite{stanley2007compositional, ha2016generating}, shape  and scene representation \cite{mescheder2019occupancy, genova2019learning, genova2020local, chen2019learning, zeng2022lion, sitzmann2019scene, jiang2020local, mildenhall2021nerf}, audio \cite{sitzmann2020implicit}, graphs \cite{grattarola2022generalised}, and data  manifolds \cite{dupont2021generative, dupontfuncta}.  

A couple of recent works~\cite{dupont2021generative, dupontfuncta}  have relied on INRs 
to generate data at any continuous coordinate (e.g., to generate images of different resolutions). 
\citet{dupontfuncta} disentangle the task of learning functions using an INR from the data generation task; and, \citet{dupont2021generative} 
proposes a \emph{hypergenerator} based on generative adversarial networks (GANs), where the parameters of its generator are the outputs of another network. 
However, they both suffer from  limitations, especially with regard to conditional generation tasks  such as image in/out-painting. The implicit nature of \citet{dupont2021generative} does not provide straightforward ways for conditional generation, and \citet{dupontfuncta} requires solving a computationally expensive numerical optimization problem to generate the modulation vector of each new data point (i.e., unseen during training).

In this paper, we propose a Variational Mixture of HyperGenerators
for learning distributions over functions, referred to as \ours. Our model relies on a mixture of hyper variational autoencoders (VAEs)
\cite{kingma2013auto, rezende2014stochastic, nguyen2021variational}, where: 
\begin{itemize}[noitemsep,topsep=0pt,parsep=0pt,partopsep=0pt]
\item[i)]  a planar normalizing flow \cite{Rezende2015VariationalIW} is used as prior distribution over its latent variables to be able to fit and generate complex data; 
\item[ii)] a hypernetwork  is used to parameterize the mixture of decoders that, in turn,   partition  the function space into meaningful regions (e.g., into visual segmentation maps as shown in~\cref{fig:segmentation}); and, 
\item[iii)]  %
analogous to \citet{dupont2021generative}, a PointConv network~\cite{wu2019pointconv} is used in the encoder to map any cloud of points, i.e., any set of continuous coordinates and features, into a fixed-sized vector.
\end{itemize}
As demonstrated by our extensive experiments on several benchmark datasets, \ours\ can accurately and efficiently learn distributions over functions and, thus,  generate data over continuous coordinate systems. 
Remarkably,  in contrast to prior work, conditional generation (e.g.,  in/out-painting tasks or generating a higher-resolution version of a given image) is straightforward in \ours\ as it just requires a forward pass on the model (independently of whether the conditioning data were seen during training). 

%% file: sections/related_work.tex
\section{Related Work \& Background}
\myparagraph{Variational Autoencoders (VAEs)} \cite{kingma2013auto, rezende2014stochastic} 
approximate the intractable posterior over latent variables $p(\zb|\xb)$ by 
performing amortized variational inference \cite{cremer2018inference, zhang2018advances}
with an auxiliary model that obtains the  approximation $q_\phi(\zb |\xb)$, 
using an encoder-decoder architecture.  

Their objective is the Evidence Lower Bound (ELBO),
\begin{align}
\mathcal{L}(\xb) = \E[\qphi] { \log \ptheta[\xb | \zb] } - \kld{\qphi[\zb | \xb]}{p(\zb)},
\end{align}
which encourages proper data reconstruction via the first term, whilst minimizing mismatch between posterior and prior via the second term.

When complex data spaces are to be encoded in the latent space, flexible priors are required to avoid a significant mismatch between the aggregated posterior and the prior, typically referred to as the \emph{prior hole problem }~\cite{rezende2018taming}. In previous works, this issue has been alleviated by various strategies, including using multimodal priors mimicking the aggregated posterior (VampPrior) \cite{tomczak2018vae}, or training flow-based \cite{Rezende2015VariationalIW, kingma2016improved, nf_prob, gatopoulos2021self}, autoregressive \cite{chen2017variational} or hierarchical priors \cite{klushyn2019learning, maaloe2019biva, peis2022missing, zeng2022lion}. However, these methods are all tailored for structured data under grid representations, whereas our proposed model is specifically designed for unstructured data.

\myparagraph{Implicit Neural Representations (INRs) } INRs represent a powerful approach for parameterizing non-linear continuous functions that map coordinates to data using deep neural networks.  This allows for efficient and independent querying of continuous locations, which is useful for various tasks like learning, graphics, vision, and graphs \cite{stanley2007compositional, ha2016generating, mescheder2019occupancy, genova2019learning, genova2020local, chen2019learning, zeng2022lion, sitzmann2019scene, jiang2020local, mildenhall2021nerf,sitzmann2020implicit,grattarola2022generalised,dupont2021generative, dupontfuncta}.

Earlier versions of INRs struggled to capture high-frequency details but advancements have addressed this issue through improved input encoding \cite{Tancik20}, activation functions, and network architectures. Recently, \citet{hao2022loe} proposes a novel INR Levels-of-Experts (LoE) model that generalizes INRs based on MLPs with position-dependent weights, greatly increasing the model capacity. On the downside, the complexity of coordinate-dependent generation functions rules out the use of hypernetwork-based approaches.

\myparagraph{HyperNetworks} HyperNetworks \cite{hypernetworksha} are a powerful class of neural networks that generate the weights for a principal network. Recently, \citet{nguyen2021variational} combined hypernetworks with VAEs in order to improve performance and generalization when modeling different tasks concurrently. In this approach, hypernetworks are used to generate the parameters of the approximate posterior ($\phib$) and the likelihood ($\thetab$).
In contrast, our proposed approach draws inspiration from the generator of \gasp\ \cite{dupont2021generative} and utilizes the hypernetwork to output the parameters  of our data generator. 

\myparagraph{Deep Generative Models for INRs}
Recently, several works have proposed to use INRs for learning distributions of functions, rather than distributions of data directly.  
\citet{dupontfuncta} propose to disentangle the task of learning functions by first learning the named \emph{functas}, or modulation vectors that configure an INR for each datapoint. 
SIREN \cite{sitzmann2020implicit} is used as the base INR network. In the second stage, any deep generative model can be trained on the learned \emph{functaset}. Then, for conditional generation tasks, computing the modulation vector for a test point requires solving a numerical optimization problem.
In \cite{rodriguez2022function}, the INR generator is constructed by combining a Bayesian NN with Gaussian weight priors that takes as input both the coordinates and a sample from a latent Gaussian noise distribution. $\alpha$-divergence Variational Inference is developed by jointly approximating the function at different coordinates using a Gaussian Process. Inducing points are introduced to make the model scalable. While the performance of the method is remarkable in a small-to-moderate dimension and it provides posterior inference, scaling it to model high-dimensional objects such as images is certainly not trivial.

In GASP \cite{dupont2021generative}, a generator of functions is built by transforming samples from a standard Gaussian latent variable to 
sets of weights using a hypernetwork. They train this generator using a GAN-style approach jointly with a PointConv-based discriminator that tries to discern fake from real samples. The LoE model in \cite{hao2022loe} is also reformulated in a generative way by taking as input both the coordinate vector and a Gaussian latent noise sample and it is also adversarially trained. 
Therefore, we can state that across \functa, \gasp\ and LoE, inference over test data to perform conditional generation, for instance, image completion or super-resolution, is not trivial as they all require numerical optimization to find the latent codes. 
In \ours,  we use a similar generator approach as in \gasp, but we generalize it by using a mixture of hypergenerators combined with a flexible latent space constructed using normalizing-flows \cite{Rezende2015VariationalIW}. 
Additionally, \ours\ can be robustly trained using stochastic variational inference and do not require any extra optimization to perform inference on unseen data. 
The proposed method is a significant advancement in the field of INRs and deep generative models, and we demonstrate it has the potential to achieve state-of-the-art results in various tasks.

%% file: sections/our_model_v3.tex
\section{Variational Mixture of HyperGenerators}

\begin{figure}[t!]
\centering
\begin{subfigure}{.5\columnwidth}
    \centering
    \includegraphics[height=\textwidth]{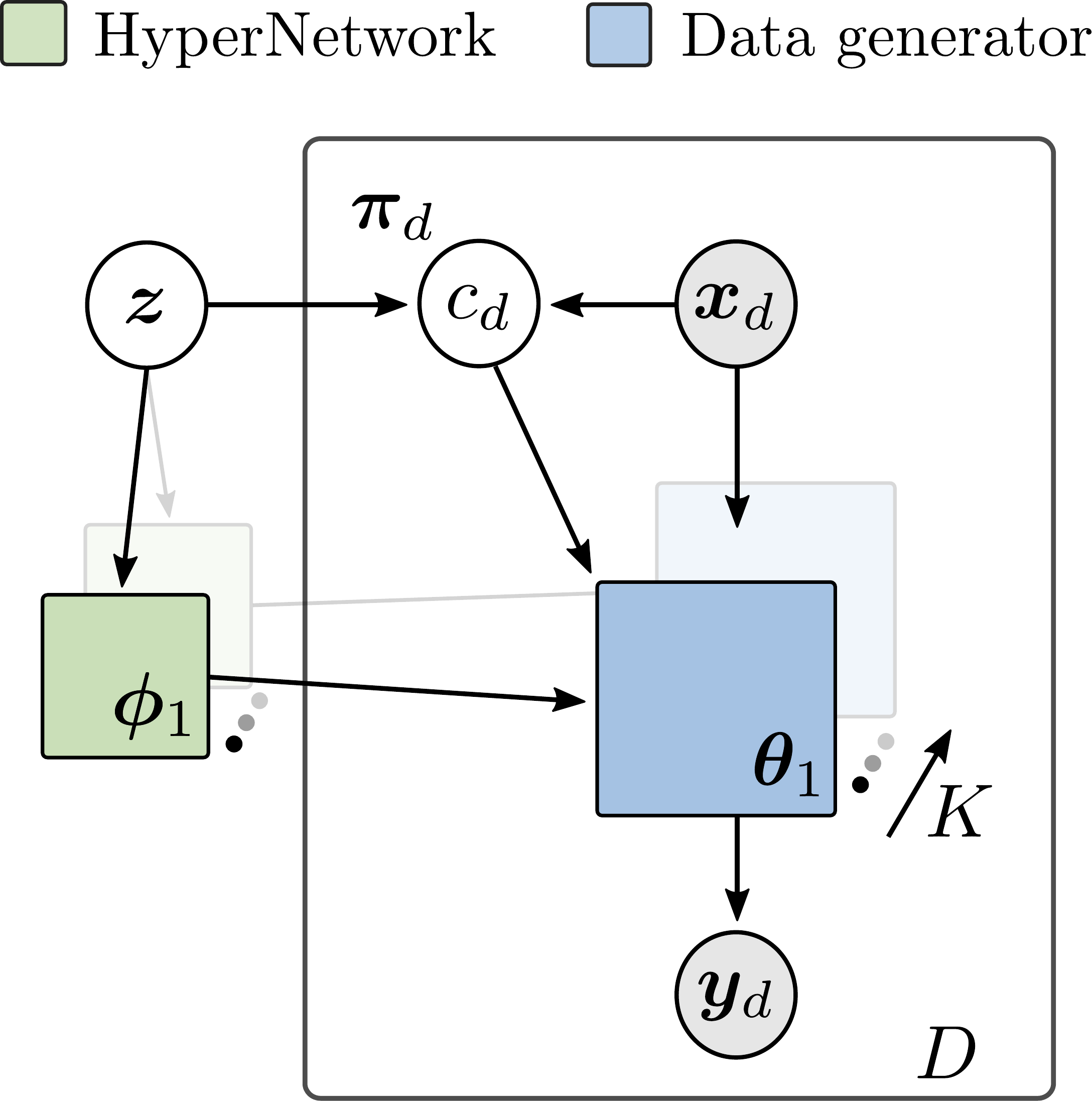}
    \caption{Generative model}\label{fig:generative}
\end{subfigure}%
\begin{subfigure}{.5\columnwidth}
    \centering
    \includegraphics[height=\textwidth]{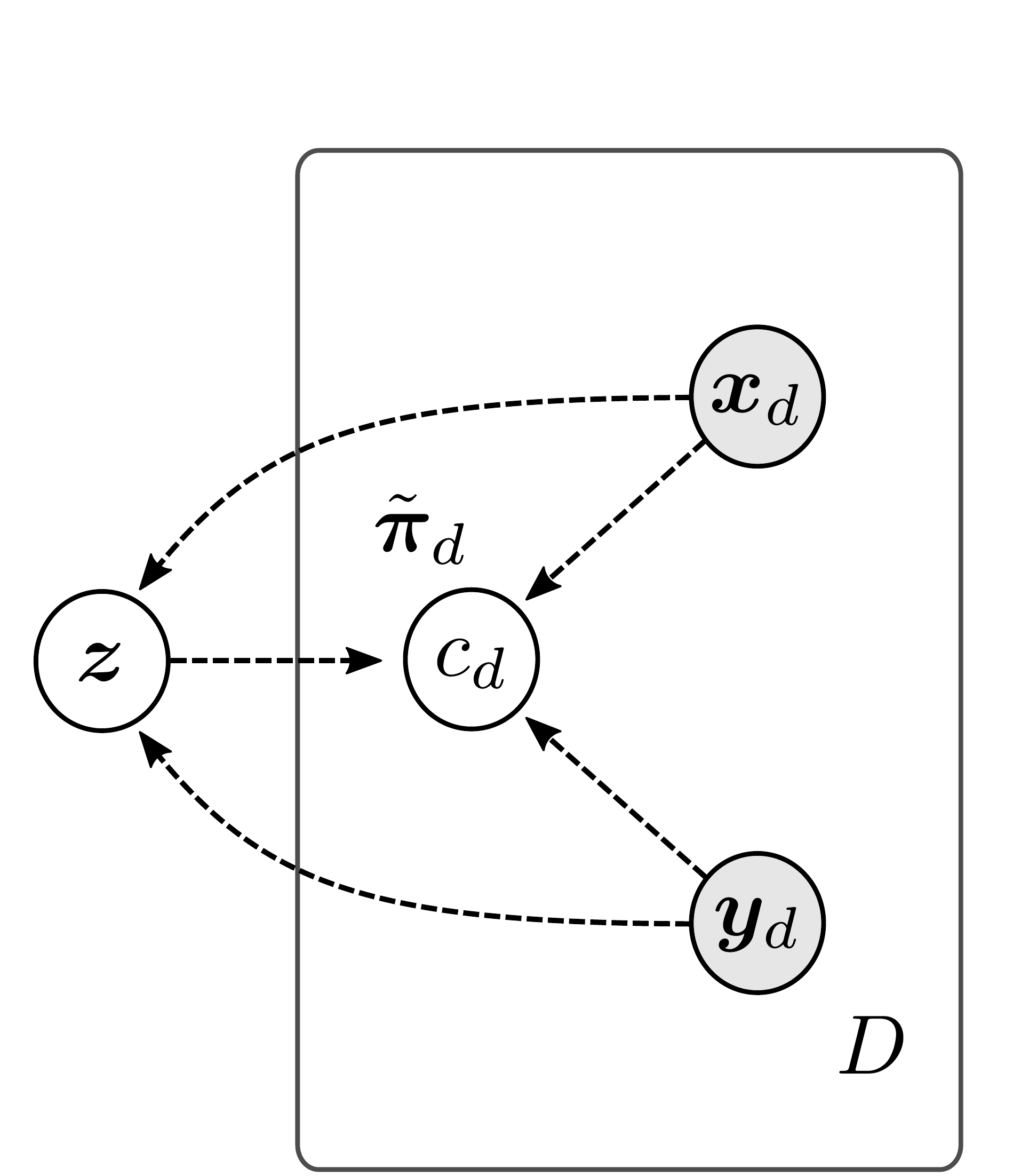}
    \caption{Inference model}\label{fig:inference}
\end{subfigure}%
\caption{The VAMoH generative (a) and inference (b) model.}
\label{fig:vamoh}
\end{figure}

In this section, we introduce the \textbf{Va}riational \textbf{M}ixture \textbf{o}f \textbf{H}yperGenerators (\ours) model. \ours\ seamlessly integrates the capabilities of Variational Autoencoders (VAEs), mixtures of generative models, and hypernetworks to handle continuous domain data points effectively. Additionally, by incorporating normalizing flows \cite{Rezende2015VariationalIW} as an expressive prior and utilizing Implicit Neural Representations (INRs) and Point Cloud encoders \cite{wu2019pointconv}, \ours\ achieves superior performance and interpretability in a variety of both sample generation and inference tasks, such as in-painting and out-painting. The proposed generative model is depicted in Figure \ref{fig:generative}.

\raggedbottom

\myparagraph{Notation} 

We denote $[D] = \{1 \dots D\} $ to the set of positive integers from 1 to D.
Let  $(\bm{X}^{(i)}, \bm{Y}^{(i)})$, $i \in [N]$ be a set of $N$ data samples (e.g., images). The $i$-th sample comprises a point cloud of $D_i$ coordinate vectors, $\bm{X}^{(i)} \doteq \{\xb^{(i)}_{j}\}_{j=1}^{D_i}$, and the set of corresponding feature vectors $\bm{Y}^{(i)}\doteq\{\bm{y}^{(i)}_j\}_{j=1}^{D_i}$. Let $\Xcal$ and $\Ycal$ denote the space of coordinate and feature vectors, respectively. As an example, in the context of image analysis, $D_{i}$ represents the number of pixels in image $i$, and $(\bm{X}^{(i)}, \bm{Y}^{(i)})$ correspond to the set of $\Rbb^2$ coordinates and values (e.g. RGB values) of the pixels, respectively.

\subsection{Mixture of HyperGenerators}
\label{subsec:generative_model}

\ours\; generates a feature set $\bm{Y}$ given a set of corresponding coordinates $\bm{X}$. For simplicity, let us assume that $(\bm{X}, \bm{Y})$ is an image with $D$ pixels.  To generate such an image, first a continuous latent variable $\zb$ is sampled from a prior distribution $p_{\psib_z}(\zb)$ parameterized by $\psib_z$. The resulting vector $\zb$  acts as  the input to $K$ different \emph{hypergenerators}. Here, we refer as a hypergenerator to  both an MLP-based hypernetwork $g_{\phi_k}(\zb)$, with input  $\zb$ that outputs a set of parameters $\bm{\theta}_{k} = g_{\phi_k}(\zb)$; and, a data generator, $f_{\theta_k}$, parametrized by the output of the hypernetwork. 
Thus,  both $\zb$ and $\bm{\theta}_{1}, \ldots,\bm{\theta}_{K}$ encode the information shared among the $D$ coordinates (e.g., pixel location) in the data (e.g., pixel values) generation process.

In order for the resulting model to be expressive and interpretable, we assume that \emph{each pixel is sampled from a mixture of $K$ hypergenerators}. 
Thus, for each pixel $d \in [D]$, we introduce a latent categorical variable $c_d\in [K]$ in order to select the  hypergenerator responsible of the pixel distribution, such that
\begin{equation}\label{eq:gen}
    \begin{gathered}
    p_{\psi_c}(\bm{C} | \bm{X}, \bm{z}) = 
    \prod_{d=1}^{D} \prod_{k=1}^{K}  \pi_{dk}^{\llbracket c_{d}=k \rrbracket},
    \end{gathered}
\end{equation}
where $\llbracket c_{d}=k \rrbracket$ is the indicator function. 
The probability mass function (pmf) of $c_d$ is also parameterized using a neural network $f_{\psi_c}$ with parameters $\psi_c$ that takes both $\xb_d$ and $\bm{z}$ as input, and outputs $\bm{\pi}_{d}=f_{\psi_c}(\xb_d,\bm{z})$ using a soft-max function, where $\pi_{dk}=P(c_d=k|\xb_d,\bm{z})$.

In summary, the overall generative process is given by:  
\begin{equation}\label{eq:gen_simple}
    p(\bm{Y}, \bm{C}, \bm{z} | \bm{X}) =
    p_{\psi_z}(\zb) \prod_{d=1}^{D} \prod_{k=1}^{K}  
\pi_{dk}^{\llbracket c_{d}=k \rrbracket} \, p_{\bm{\theta_k}}(\bm{y}_{d}|\xb_d),
\end{equation}
where, importantly, we rely on Random Fourier Features (RFF) \cite{Tancik20} to encode $\xb_d$ to capture high frequency details with our generative functions $f_{\bm{\theta}}$, as  in \citet{dupont2021generative}.

\begin{figure}[t]
\vspace{0.5cm}
    \centering
    \centerline{\includegraphics[width=\columnwidth]{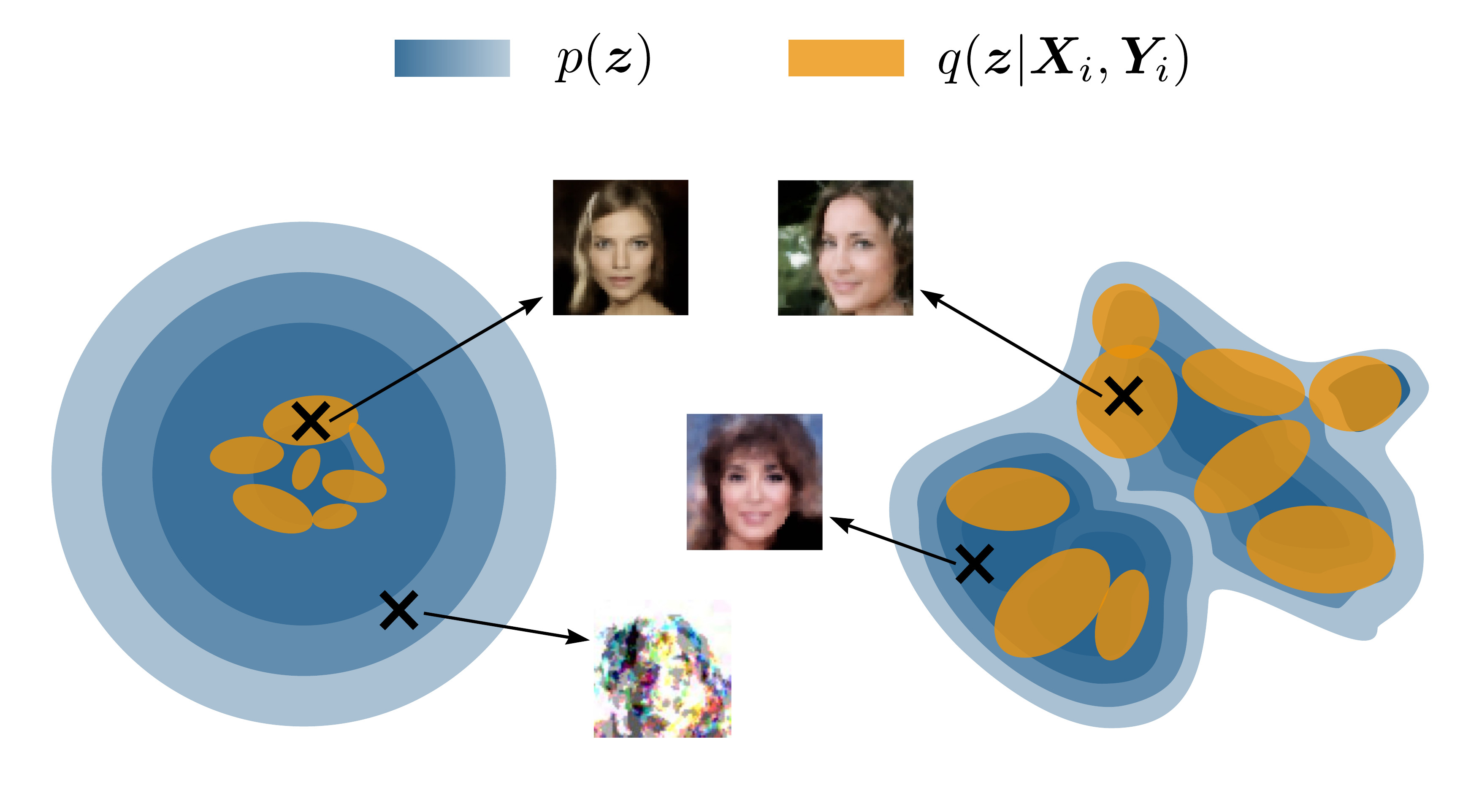}}
    \caption{Illustration of the \emph{prior hole problem}. Blue contours are the prior $p(\zb)$, whilst orange contours are approximate posteriors $q(\zb |\bm{X}_i, \bm{Y}_i)$. Left: simple standard prior does not accurately cover the encoder complexity. Decoding samples from the prior that fall far from the aggregated posterior from training data gives unrealistic images. Right: a more flexible prior properly matches the complexity of the encoder, leading to better quality of the images generated from the prior.}
    \label{fig:hole}
\end{figure}
\raggedbottom

\myparagraph{Flow-based prior on $\zb$}

The use of a fixed prior distribution for generating new unconditional samples in VAEs often results in the well-documented \emph{prior hole problem }~\cite{rezende2018taming},  illustrated in Figure \ref{fig:hole}. This limitation stems from the poor expressiveness of a fixed prior distribution as compared to the approximate posterior. To tackle this issue, we rely on %
normalizing flows (NF) to learn a prior distribution of the  form $p_{\psi_z}(\zb)$ with parameters $\psi_z$. This improves the prior expressiveness and,  thus, addresses the aforementioned problem. Specifically, \ours\; integrates $T$  %
layers of {a}  planar flow~\cite{Rezende2015VariationalIW} with a Gaussian $p_0(\zb)$ as base distribution, resulting in a flexible prior distribution that significantly enhances  the generated samples. Extended empirical analysis is provided in Appendix \ref{app:flow_prior}.

\input{sections/our_model_inference_short.tex}
\subsection{Training}\label{sec:training}
The evidence lower bound (ELBO) of our proposed model is given by
\begin{align}
\label{eq:elbo_ours}
\mathcal{L}(& \bm{X},  \bm{Y})  = \E[q_{\gamma}] { \log \ptheta[ \bm{Y}| \zb, \bm{X}, \bm{C}] } \nonumber\\
& - \kld{q_{\gamma_{z}}(\zb | \bm{X}, \bm{Y} )}{p_{\psi_{z}}(\zb)} \\
 & - \E[q_{\gamma_{z}}] { \kld{q_{\gamma_{c}}(\bm{C} |\zb, \bm{X}, \bm{Y} )}{p_{\psi_{c}}(\bm{C} | \zb,  \bm{X})}}. \nonumber
\end{align}

We provide the complete derivation in \cref{app_sec:elbo}. We train \ours\ maximizing this ELBO by stochastic gradient descent on randomly selected mini-batches. To stabilize the initialization of the normalizing flow and accelerate convergence, we employ a standard prior $p(\zb) = \mathcal{N}(\bm{0}, \bm{I})$ during the initial epochs of the training to allow the model to focus on organizing the approximate posterior and producing accurate reconstructions. After some iterations, we start training the Planar Flow $p_{\psi_z}(\zb)$.
While KL between discrete distributions in \cref{eq:elbo_ours} is easily computed in closed form, the continuous KL is approximated by Monte Carlo sampling, just as we do for the reconstruction term. 

\begin{table*}[h!]
    \centering
\setlength\tabcolsep{3pt}
    \centering
    \caption{Comparison of FID and Precision and Recall scores of image generation for \ours, \gasp, and \functa.  Low FID, high precision, and high recall indicate the best performance. The best results are highlighted in bold.  Note that  for \functa\ and \celebahq\ we just report their FID value, since they do not report the precision and recall.}
    \label{table:comparison}
    \vskip 0.15in
    \begin{tabular}{c ccc ccc}
\toprule
 &  \multicolumn{3}{c}{\celebahq}  &  \multicolumn{3}{c}{\shapesd}  \\
   \cmidrule(r){2-4}   \cmidrule(r){5-7}
  Model  &  $\downarrow$ FID &  $\uparrow$ Precision  &  $\uparrow$ Recall  &  $\downarrow$ FID &  $\uparrow$ Precision  &  $\uparrow$ Recall     \\
  \cmidrule(r){1-7} 
    \gasp\ \cite{dupont2021generative} & \bfseries 14.01 $\pm$  0.18 & \bfseries 0.81 $\pm$ 0.0 & \bfseries 0.43 $\pm$ 0.01 & 118.66 $\pm$ 0.64 & 0.01 $\pm$ 0.0 & 0.16 $\pm$ 0.01 \\
      \functa\ \cite{dupontfuncta} &  40.40  &  -  &  -  & 57.81 $\pm$ 0.15 & 0.06 $\pm$ 0.0 & 0.13 $\pm$ 0.0 \\
  \ours  &66.27 $\pm$ 0.18 & 0.65 $\pm$ 0.0 & 0.0 $\pm$ 0.0 & \bfseries 56.25 $\pm$ 0.57 & \bfseries 0.08 $\pm$ 0.0 & \bfseries 0.64 $\pm$ 0.01 \\
\bottomrule
\end{tabular}
\vspace{-5pt}
\end{table*}

\myparagraph{Point dropout}
The design of VAMoH allows for easily handling point clouds $\{\bm{X}, \bm{Y}\}$ of arbitrary sizes. The PointConv encoder is able to convolve the observed points, regardless of their coordinates, and map them into the approximate posterior, which can then be decoded to generate new points at any desired location. To enhance the conditional generation capabilities and robustness of VAMoH in inferring information from partial data, we apply dropout to the points within a set $\{\bm{X}, \bm{Y}\}$ according to a probability $p\sim U(0, \alpha)$, which is sampled independently for each batch. The maximum dropout probability, $\alpha$, is fixed for ensuring that the reduced set contains at least as many points as centroids to be found at the first layer of PointConv. In previous VAE-based models \cite{ma2020vaem, peis2022missing}, this strategy has been successfully employed for masking training batches in order to improve missing data imputation tasks. Nevertheless, since these models deal with grid-type incomplete data, some pre-imputation is required before feeding the encoder, mean imputation or zero filling being typical choices, with the cost of introducing bias in the model \cite{simkus2021variational}. In contrast, in our work, the PointConv encoder easily handles missing data, without requiring any pre-imputation strategy. Additional empirical evaluations can be found in the Appendix \ref{app:point_dropout}.

%% file: sections/our_model_inference_short.tex
\subsection{Inference model}

In this subsection, we present the inference model that we propose to approximate the posterior of the latent variables $\zb$ and $\bm{C}$. The model is defined as follows:

\begin{equation}
     q_{\gammab}(\zb, \bm{C}|\bm{Y}, \bm{X}) 
 = q_{\gammabz}(\zb|\bm{Y}, \bm{X}) \prod_{d=1}^{D} q_{\gammabc}(c_{d}|\zb,\bm{y}_{d}, \xb_{d}).
\end{equation}

We illustrate the inference model in Figure~\ref{fig:inference}. Throughout the following, we refer to all the parameters of the inference model as $\gammab=\{ \gammabz, \gammabc \}$.

\myparagraph{Inference for the continuous latent variable} We propose to model the posterior distribution of $\zb$  as $q_{\gammabz}(\zb|\bm{Y}, \bm{X}) = \normal{\zb | f_{\gammabz} (\bm{Y}, \bm{X}) }$, parameterized by $\gammabz$.
It's important to note that this distribution is shared among the  complete sample (e.g., image), thus  $\zb$ contains global information.
To parametrize $f_{\gammabz}$, we use a PointConv \cite{wu2019pointconv}. This entails several advantages. Firstly, it generalizes convolutional operations to continuous space coordinate systems, in contrast to the fixed grids used in CNNs \cite{lecun1995convolutional}. 
Secondly, it is independent of data resolution, geometry of the grid, and missingness of the data. Of particular interest is the latter, as grid-based architectures typically require missing dimensions to be filled (e.g., with zeros), which introduces bias to the model \cite{simkus2021variational}.

\myparagraph{Inference of the mixture components}
\label{subsec:inference_model}

We model the posterior over the categorical latent variables $c_{d} $ of each coordinate $d$  as  

\begin{equation}q_{\gammabc}(c_{d}| \cdot) = \cat{c_{d}| f_{\gammabc}(\zb,\bm{y}_{d}, \xb_{d})} =  \prod_{k=1}^{K} \tilde{\pi}_{dk}^{ \indic{c_{d}=k}},
\end{equation}
where we parametrize $\gammabc$ with  a simple MLP. It is important to notice that this posterior depends on the local information of the coordinate, i.e., $(\xb_{d},\bm{y}_d)$, and only on the global information through $\zb$. This choice is made to encourage the model to learn to use different hypergenerators for different parts of the data.

%% file: sections/experiments.tex
\begin{figure}[t]
\begin{center}
\raisebox{10pt}{\rotatebox[]{90}{\small \functa}}\hspace{5pt}
\includegraphics[width=0.93\columnwidth]{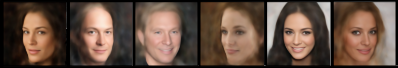}
\raisebox{10pt}{\rotatebox[]{90}{\small \gasp}}\hspace{5pt}
\includegraphics[width=0.93\columnwidth]{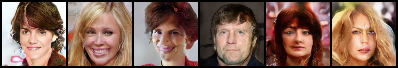}
\vspace{10pt}
\raisebox{12pt}{\rotatebox[]{90}{\small \ours}}\hspace{5pt}
\includegraphics[width=0.93\columnwidth]{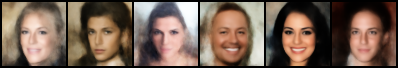}
\raisebox{10pt}{\rotatebox[]{90}{\small \functa}}\hspace{5pt}
\includegraphics[width=0.93\columnwidth]{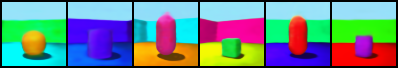}
\raisebox{10pt}{\rotatebox[]{90}{\small \gasp}}\hspace{5pt}
\includegraphics[width=0.93\columnwidth]
{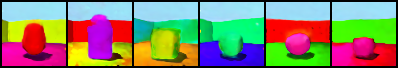}
\raisebox{12pt}{\rotatebox[]{90}{\small \ours}}\hspace{5pt}
\includegraphics[width=0.93\columnwidth]{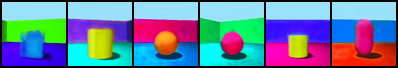}
\end{center}
\caption{Comparison of generation quality at original resolution. \celebahq\ samples of \functa\ obtained from \cite{dupontfuncta}. }
\label{fig:generation}
\end{figure}
\raggedbottom

\begin{figure}[t]
\centering
\raisebox{10pt}{\rotatebox[]{90}{\small \gasp}}\hspace{5pt}
\includegraphics[width=0.93\columnwidth]{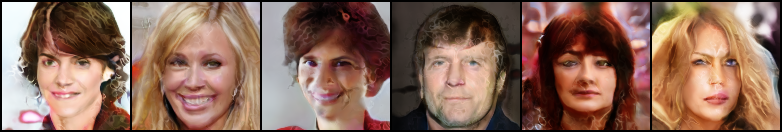}
\raisebox{12pt}{\rotatebox[]{90}{\small \ours}}\hspace{5pt}
\vspace{10pt}
\includegraphics[width=0.93\columnwidth]{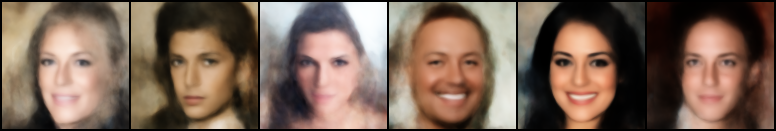}
\raisebox{10pt}{\rotatebox[]{90}{\small \gasp}}\hspace{5pt}
\includegraphics[width=0.93\columnwidth]{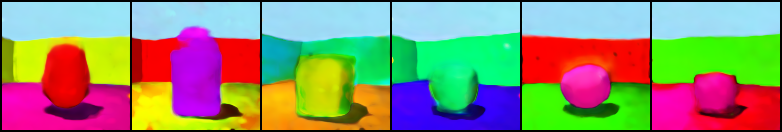}
\vspace{10pt}
\raisebox{12pt}{\rotatebox[]{90}{\small \ours}}\hspace{5pt}
\includegraphics[width=0.93\columnwidth]{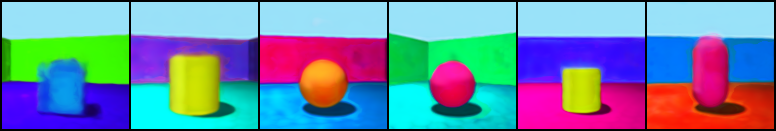}
\raisebox{20pt}{\rotatebox[]{90}{\small \gasp}}\hspace{5pt}
\includegraphics[width=0.9\columnwidth]  {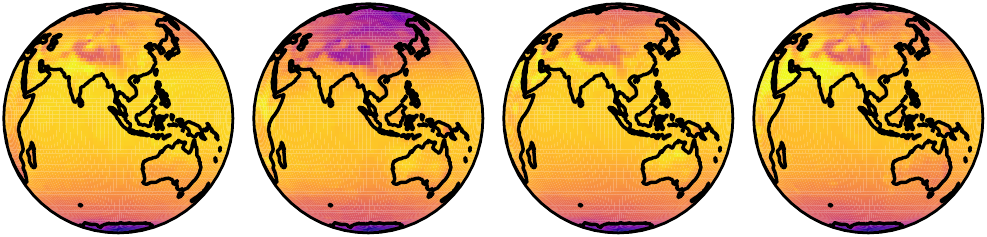}\hspace{5pt}
\raisebox{20pt}{\rotatebox[]{90}{\small \ours}}\hspace{5pt}
\includegraphics[width=0.9\columnwidth]{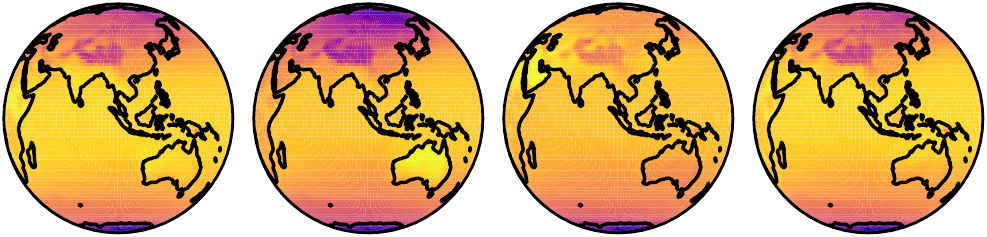}
\caption{Comparison of uncurated generated samples at super-resolution.}
\label{fig:super_generation}
\end{figure}
\raggedbottom

\section{Experiments}

In this section, we provide a thorough empirical evaluation of \ours. 
We evaluate our model on the tasks of data generation, reconstruction, and imputation, including the superresolution results. 

\myparagraph{Baselines} We compare \ours\ with \gasp\ \cite{dupont2021generative} and \functa\ \cite{dupontfuncta} with a Normalizing Flow as generator of modulation vectors, which we denote simply by \functa. More specifically, \gasp\ is only considered for data generation, since it does not allow for any inference-related task. 

\myparagraph{Datasets} We evaluate \ours\ on  \polymnist\ ($28 \times $28), \celebahq\ ($64 \times $64) \citep{karras2017progressive}, \shapesd\ ($64 \times $64) \citep{3dshapes18}, climate data from the \era\ dataset \cite{era5}, and 3D chair voxels from the \chairs\ dataset \cite{chang2015shapenet}. 
The architecture and hyperparameters used for each dataset are detailed in \cref{app:exp_setup}.

We implemented \ours\ in PyTorch and performed all experiments on a single V100 with 32GB of RAM. 
The code with the model implementation and experiments is available at 
\url{https://github.com/bkoyuncu/vamoh}. In this section, we provide a glimpse into the results of our experiments on all datasets. Additional experiments and their outcomes can be found at Appendix \ref{app:exp_ext}.
\subsection{Generation}
\label{subsec:generation}
In this section, we evaluate \ours\ for the task of synthetic data generation over continuous coordinate systems. Samples are obtained from the learned models in both the original and twice the resolution, referred to as \textit{super resolution}. 
\myparagraph{Metrics}  We use two metrics to evaluate the quality of the generated samples. We report the usually employed Fréchet Inception Distance (FID) \cite{heusel2017gans} as well as the \textit{improved precision and recall} \cite{kynkaanniemi2019improved}, which measures the quality of the generated data (i.e, high precision) and the coverage of the true data distribtuion (i..e, high recall).
\myparagraph{Results} 
We present the summary of the quantitative comparison of \ours\ with both baselines for two image datasets in \cref{table:comparison}.
Our analysis indicates that, quantitatively speaking, there is not a clear winner among the models.
While \gasp\ achieves the highest performance
on the \celebahq\ dataset,  visual inspection of the generated samples in \cref{fig:generation} reveals that the quality of \functa\  and \ours\ is also high.
This aligns with previous studies \cite{dupontfuncta} that have noted that FID may over-penalize blurriness.
Additionally, it is worth noting that \ours\ obtains a recall of 0.0 on this dataset, despite the generated samples displaying diversity in features such as facial expressions and hairstyles, as seen in \cref{fig:generation}.
In analyzing the performance of \ours\ on the \shapesd\ dataset, we observe a clear superiority compared to other models.
As depicted in \cref{fig:generation}, \ours\ is able to generate objects with diverse shapes and colors, as well as  walls and floors delimited with sharp edges. 
Furthermore, it achieves the highest quantitative metrics, as demonstrated in the right columns of \cref{table:comparison}. 
Despite this, it is important to note that the low precision values obtained by all models require a comprehensive evaluation approach, incorporating both quantitative and qualitative metrics when assessing generative models.

Finally, in \cref{fig:super_generation}, we evaluate the super-resolution capabilities of the models by visually inspecting the generated samples at twice the original resolution. 
We utilize the same latent code as in the generated samples in \cref{fig:generation} for the \celebahq\ and \shapesd\ datasets. 
Additionally, to demonstrate the versatility of our \ours\ to model diverse types of data, we also include super resolution samples from the \era\ dataset.
Comparison with super-resolution samples generated by \functa, and additional results for all the datasets can be found in \cref{app:generation}.

\myparagraph{Discussion}
A significant advantage of \ours\ is its ability to achieve the capability of generation through a single optimization procedure. In contrast, \gasp\ requires solving the min-max GAN optimization, which has been acknowledged to be unstable \cite{Jabbar2020ASO}; and \functa\ requires first learning the SIREN model and modulations for each sample, followed by the training of an additional generative model, such as a normalizing flow.

Overall, these results demonstrate that the generation quality of \ours\ is comparable to existing alternatives.
In the following, we show that \ours\ exhibits superior efficiency and performance in inference tasks in comparison with \functa. 
It is worth noting that we do not compare \ours\ with \gasp, as it is purely a generative model.

\begin{figure*}[t!]
\vspace{0.5cm}
\begin{subfigure}{.47\textwidth}
    \centering
    Ground truth
    \begin{flushright}
    \includegraphics[width=0.9\columnwidth]{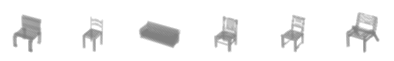}
    \end{flushright}
    Reconstructions
    \begin{subfigure}{0.9\columnwidth}
    \raisebox{15pt}{\rotatebox[]{90}{\small\functa}}\hspace{5pt}
    \includegraphics[width=\columnwidth]{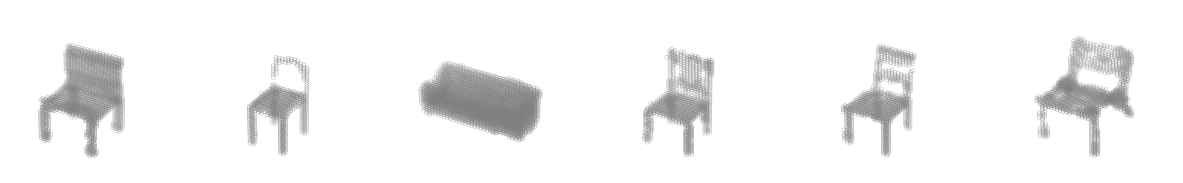}
    \raisebox{12pt}{\rotatebox[]{90}{\small\ours}}\hspace{5pt}
        \includegraphics[width=\columnwidth]{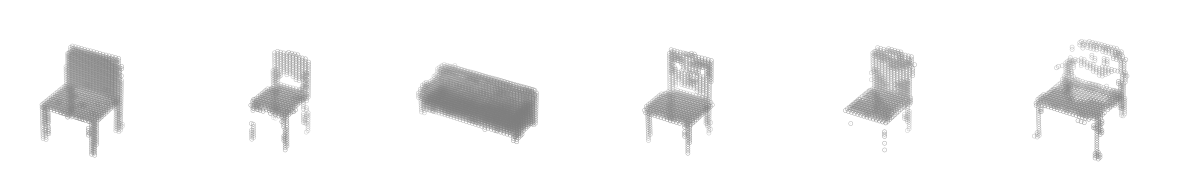}
    \end{subfigure}%
    \\
    Super-reconstructions
    \begin{subfigure}{0.9\columnwidth}
    \raisebox{15pt}{\rotatebox[]{90}{\small\functa}}\hspace{5pt}
     \includegraphics[width=\columnwidth]{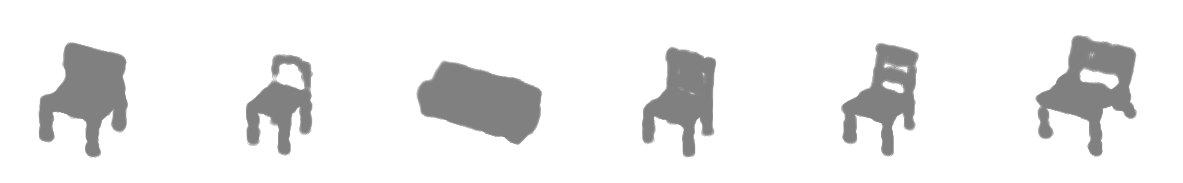}
     \raisebox{12pt}{\rotatebox[]{90}{\small\ours}}\hspace{5pt}
        \includegraphics[width=\columnwidth]{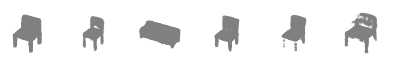}
    \end{subfigure}
    \caption{\chairs}
    \label{fig:recons_shapenet}
\end{subfigure}%
\hfill
\begin{subfigure}{.47\textwidth}
    \centering
    Ground truth
    \begin{flushright}
    \includegraphics[width=0.9\columnwidth]{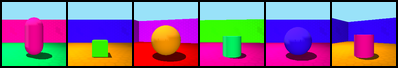}
    \end{flushright}
    \vspace{2pt}
    Reconstructions
    \begin{subfigure}{0.9\columnwidth}
    \raisebox{15pt}{\rotatebox[]{90}{\small\functa}}\hspace{5pt}
        \includegraphics[width=\columnwidth]{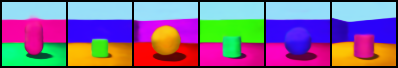}
        \vspace{5pt}
        \raisebox{12pt}{\rotatebox[]{90}{\small\ours}}\hspace{5pt}
        \includegraphics[width=\columnwidth]{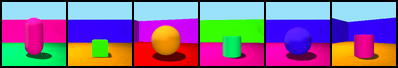}
    \end{subfigure}%
    \vspace{2pt}
    Super-reconstructions
    \begin{subfigure}{0.9\columnwidth}
    \raisebox{15pt}{\rotatebox[]{90}{\small\functa}}\hspace{5pt}
        \includegraphics[width=\columnwidth]{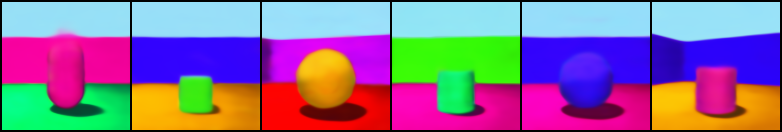}
        \raisebox{12pt}{\rotatebox[]{90}{\small\ours}}\hspace{5pt}
        \includegraphics[width=\columnwidth]{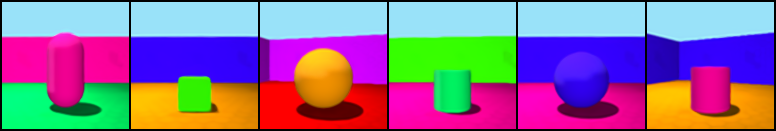}
    \end{subfigure}
    \caption{\shapesd}
    \label{fig:recons_shapes3d}
\end{subfigure}%
\caption{Comparison of reconstruction quality of \ours\ and \functa\ on ground truth data from the first samples of the test set at original and super-resolution.}
\label{fig:recons}
\end{figure*}
\raggedbottom

\subsection{Reconstruction}

\begin{table*}[ht!]
    \centering
\setlength\tabcolsep{3pt}
    \centering
    \caption{Comparison of inference time (seconds)  for reconstruction task of \ours\ and \functa. On the right-most two columns, we show the speed improvement of VaMoH compared to Functa (3) which is trained with 3 gradient steps as suggested in the original paper \cite{dupontfuncta}
    and Functa (10) which is trained with 10 gradient steps to obtain the results of Functa depicted in Figures \ref{figapp:recons_celeba},\ref{figapp:recons}. Please note that these experiments are run on the same GPU device.}
    \label{table:time_recons}
    \vskip 0.15in
    \begin{tabular}{c ccc cc}
\toprule
 &  \multicolumn{3}{c}{Model Inference Time (secs)}  &  \multicolumn{2}{c}{Speed Improvement}  \\
   \cmidrule(r){2-4}   \cmidrule(r){5-6}
  Dataset  &  \ours\ &  \functa\ (3) &  \functa\ (10)  &  vs. \functa\ (3)  & vs. \functa\ (10)    \\
  \cmidrule(r){1-6} 
  \polymnist  &\bfseries0.00453 & 0.01648 & 0.05108 &\bfseries x 3.64 &\bfseries x 11.28\\
  \shapesd  &\bfseries 0.00536 & 0.01759 &	0.05480 	&\bfseries x 3.28 &\bfseries x 10.22 \\
  \celebahq  &\bfseries 0.00757 &0.01733 &0.05381 &\bfseries x 2.29 &\bfseries x 7.11\\
  \era  &\bfseries 0.00745 & 0.01899 &	0.05932 &\bfseries x 2.55 	&\bfseries x 7.96\\
  \chairs  &\bfseries 0.00689 	& 0.02095 &0.06576 & \bfseries x 3.04 & \bfseries x 9.54 \\
\bottomrule
\end{tabular}
\vspace{-5pt}
\end{table*}
\label{subsec:reconstruction}
In this section, we evaluate the performance of \ours\ in reconstructing data and compare it with \functa. As before, we conduct the assessment at both the original resolution and at double the resolution, which we refer to as \textit{super-reconstruction}. This latter scenario involves generating features at new coordinate positions.

\myparagraph{Metrics} We use the Peak Signal-to-Noise-Ratio (PSNR) to quantify the quality of reconstructions (See \cref{app:reconstruction} 
 for further details). We also compare inference times (in seconds) of \ours\ and \functa\ in Tables \ref{table:time_recons} and \ref{table:time_superrecons}.

\begin{figure}[h!]
\begin{center}
\centerline{\includegraphics[width=0.9\columnwidth]{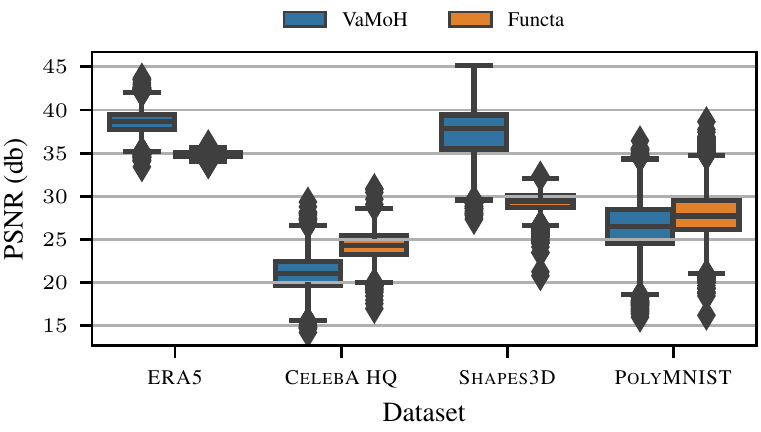}}
\caption{Comparison of PSNR (db) of reconstructed images in the test set.}
\label{fig:recons_psnr}
\end{center}
\vskip -0.2in
\end{figure}
\raggedbottom

\begin{figure}[h!]
\centering
Ground truth
\includegraphics[width=0.93\columnwidth]{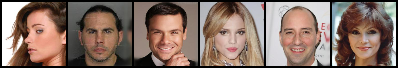}
Super-reconstruction
\includegraphics[width=0.93\columnwidth]{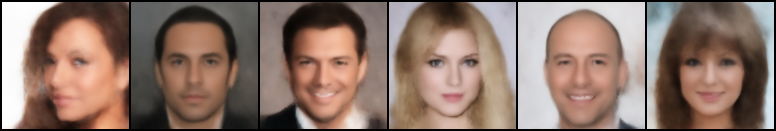}
Ground truth
\includegraphics[width=0.92\columnwidth]{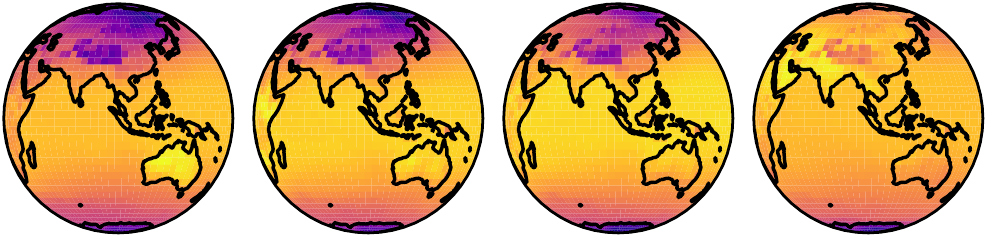}
Super-reconstruction
\includegraphics[width=0.93\columnwidth]{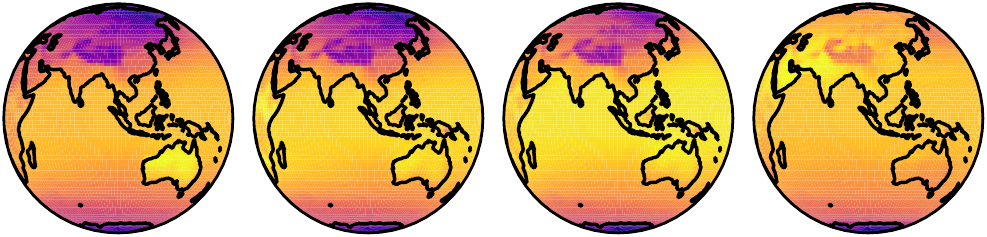}
\caption{Ground truth for the first six test set samples and the corresponding super-reconstruction obtained by \ours\ for \celebahq\ and \era\ datasets. }
\label{fig:recons_super}
\end{figure}
\raggedbottom

\begin{figure}[t]
\begin{subfigure}{.47\textwidth}
    \centering
    \raisebox{15pt}{\rotatebox[]{90}{In}}\hspace{5pt}
    \includegraphics[width=0.93\columnwidth]{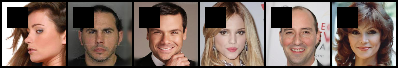}
    \vspace{10pt}
    \raisebox{12pt}{\rotatebox[]{90}{Recons.}}\hspace{5pt}
    \includegraphics[width=0.93\columnwidth]{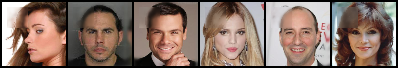}
    \raisebox{10pt}{\rotatebox[]{90}{In}}\hspace{5pt}
    \includegraphics[width=0.93\columnwidth]{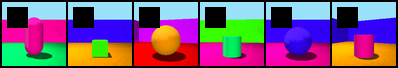}
    \raisebox{12pt}{\rotatebox[]{90}{Recons.}}\hspace{5pt}
    \includegraphics[width=0.93\columnwidth]{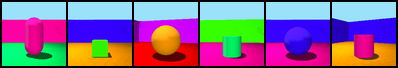}
    \caption{Missing a patch}
    \label{fig:completation_left}
\end{subfigure}%
\hfill
\vspace{2pt}
\begin{subfigure}{.47\textwidth}
    \centering
    \raisebox{15pt}{\rotatebox[]{90}{In}}\hspace{5pt}
    \includegraphics[width=0.93\columnwidth]{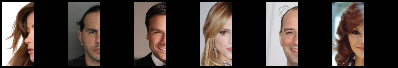}
    \vspace{10pt}
    \raisebox{12pt}{\rotatebox[]{90}{Recons.}}\hspace{5pt}
    \includegraphics[width=0.93\columnwidth]{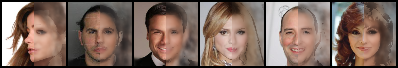} 
    \raisebox{10pt}{\rotatebox[]{90}{In}}\hspace{5pt}
    \includegraphics[width=0.93\columnwidth]{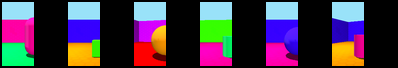}
    \raisebox{12pt}{\rotatebox[]{90}{Recons.}}\hspace{5pt}
    \includegraphics[width=0.93\columnwidth]{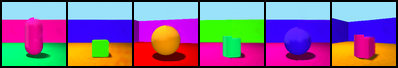}
    \caption{Missing half of the image }
    \label{fig:completation_right}
\end{subfigure}%
\caption{Imputation of different amounts of missing parts using \ours. For each dataset, the top row shows the input with the missing part (In) and the bottom row shows the reconstructed image. }
\label{fig:completion}
\end{figure}
\raggedbottom

\myparagraph{Results}
\cref{fig:recons} shows reconstructions and super-reconstructions for the first 6 samples of the test set of \chairs\ and \shapesd\ for \functa\ and \ours. 
We observe that both methods produce high-quality reconstructions.
When evaluating the \chairs\ dataset (as shown in \cref{fig:recons_shapenet}), we observe that both models are capable of capturing the details of the chairs, such as the various patterns present on the back of the chairs.
Notably, \ours\ achieves more detailed edges in both reconstructions and super-reconstructions for the \shapesd\ dataset (see \cref{fig:recons_shapes3d}) as well as overall less blurriness.
A more quantitative assessment is provided in \cref{fig:recons_psnr}, which compares PSNR values obtained for all the test set samples of the image-like datasets. 
We observe that \ours\ achieves similar quality on \polymnist\ and \celebahq, and outperforms \functa\ on \era\ and \shapesd.
This is quite remarkable as \functa\ requires solving an optimization problem per sample, and thus it `overfits' each of the samples. In stark contrast, \ours\ efficiently generates reconstructions with a single forward pass. 
As a consequence, \ours\ is significantly faster. In Table~\ref{table:time_recons}, we show that \ours\ is more than 2 times faster than \functa\ when trained with 3 gradient steps (as stated in \citep{dupontfuncta}); and at least 7 times faster when trained with 10 gradient steps. 
Finally, \cref{fig:recons_super} shows the ability of \ours\ to generate  quality super-reconstructions on the \celebahq\ and \era\ datasets. Another inference time comparison for super-reconstruction can be found in Table \ref{table:time_superrecons} in Appendix \ref{app:reconstruction}.

\myparagraph{Discussion}
Both \ours\ and \functa\ are able to reconstruct and super-reconstruct data with high quality with different data modalities (e.g., images and voxels). On top of this, \ours\ achieves comparable or better PSNR values for the test set, which indicates it may reconstruct more details.
Additionally, in Table \ref{table:time_recons}, we show that \ours\ is significantly faster than \functa\ for the reconstruction task. This happens because \ours\ requires a simple forward pass. 
In contrast, \functa\ requires finding a new modulation for each new sample, which is an optimization in itself; therefore, its inference time highly  depends on the number of gradient steps needed, which does not only have a major impact on the quality of the results but also on its computational efficiency. 
Note that this computational burden applies to any inference-related task, e.g., image completion.
Given the quantitative and qualitative assessment of the results of \ours, we argue that it might be preferable to rely on \ours\ due to its inference time efficiency in comparison to \functa. Thus, in the following, we keep understanding the capabilities of \ours\ without comparison.

\subsection{Image completion}
We evaluate the performance of \ours\ on the task of image completion.
We consider two scenarios: (i) missing a patch (i.e., image in-painting) and (ii) missing half of the image, as a challenging case. 
\cref{fig:completion} illustrates the results on unseen test samples from \celebahq\ and \shapesd. 
Our results demonstrate that \ours\ can reconstruct high-quality images, even when half of the image is missing, as in \cref{fig:completation_right}.
Thanks to the use of PointConv \cite{wu2019pointconv} as the encoder, \ours\ can infer missing information without the need of inputting dummy values (e.g., zeros) for the features of the missing coordinates. This is something needed in standard VAEs, and allows \ours\ to mitigate any potential biases in the results.
Complete results for all datasets evaluated, including those for image out-painting, can be found in \cref{app:im_completion}.

\subsection{The flexibility of VAMoH}

In this Section, we provide evidence of the  interpretability of \ours. In particular, we show an example in which the mixture of hypergenerators, with $K=4$, allows for splitting the generation of each pixel into several modes. 

\begin{figure}[t]
\centering
\raisebox{10pt}{\rotatebox[]{90}{Recons.}}\hspace{5pt}
\includegraphics[width=0.9\columnwidth]{figures/shapes3d/test/vamoh/vamoh_recons_x_mean_shapes3d.png}
\vspace{2pt}
\raisebox{10pt}{\rotatebox[]{90}{\small{$\bm{C}_{\text{ent}}$}}}\hspace{5pt}
\includegraphics[width=0.91\columnwidth]{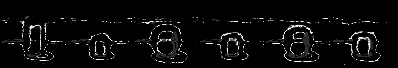}
\raisebox{10pt}{\rotatebox[]{90}{$\bm{C}_{\text{map}}$}}\hspace{5pt}
\includegraphics[width=0.9\columnwidth]{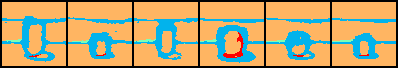}
\caption{Reconstructed samples (top), entropy (middle) $\bm{C}_{\text{ent}} = \Hcal ( \bm{C}|\bm{Y}, \bm{X} )$, and map (bottom) $\bm{C}_{\text{map}}= \max_{c_d}  q_{\gammabc}(\bm{C}|\bm{Y}, \bm{X})$ of the posterior distribution over the mixture. }
\label{fig:segmentation}
\end{figure}

In \cref{fig:segmentation}, we show the entropy (middle row) and map (bottom row) of the posterior distribution of the categorical latent variables $\bm{C}$. By examining the entropies per pixel, we observe low values for most pixels, but higher values for the borders. This indicates that the model is uncertain about which components to use at the borders, while it is using fewer components for the rest of the image. Furthermore, the map values of $\bm{C}$ resemble a segmentation map. Here,  we observe that one hypergenerator (orange) is specialized in filling the colors of different parts of the image, while the others (blue, red, and green) are used for the borders.

\raggedbottom

%% file: sections/conclusions.tex
\section{Conclusion}
In this paper, we introduced \ours, a novel VAE-based model for learning distributions of functions that enables efficient and accurate generation of data over continuous coordinate systems.
Notably, \ours\ allows for straightforward conditional generation with a simple forward pass on the model. \ours\ can perform tasks such as in-painting and out-painting or generate higher-resolution versions of a test set sample (that is, unseen during training) more efficiently than competing methods.
Our experimental results demonstrate the effectiveness of \ours, both in generation and inference tasks, in a wide range of applications and datasets.

Although our model has a couple of limitations, they are not significant factors that would impede its effectiveness. Firstly, one limitation relates to the number of parameters in the hypernet, which scales linearly with the number of components (i.e., $K$). Secondly, we could improve the speed of the PointConv encoder with a sampling-based algorithm for the selection of centroid points used for the convolution step.

As a future research direction, we aim to enhance \ours\ by exploring methods for achieving disentanglement in the latent space.
This would allow for controlled modifications of generated and reconstructed samples and may open up new possibilities for applications such as data editing at various resolutions. 
We have not identified any social concerns associated with this work. In fact, \ours\ reduces the time required for inference during testing, which could broaden the range of applications and areas where INR methods can be utilized.

%% file: sections/acknow.tex
\section{Acknowledgements}

Batuhan Koyuncu and Isabel Valera acknowledge the support by the German Federal Ministry of Education and Research (BMBF) through the Cell-o project. Pablo S\'anchez Mart\'in thanks the German Research Foundation through the Cluster of Excellence “Machine Learning – New Perspectives for Science”, EXC 2064/1, project number 390727645 for generous funding support. 
The authors thank the International Max Planck Research School for Intelligent Systems (IMPRS-IS) for supporting Pablo S\'anchez Mart\'in.
Pablo M. Olmos and Ignacio Peis acknowledge the support by the Spanish government MCIN/AEI/10.13039/501100011033/ FEDER, UE, under grant PID2021-123182OB-I00, by Comunidad de Madrid under grant IND2022/TIC-23550, and by Comunidad de Madrid and FEDER through IntCARE-CM. The work of Ignacio Peis has been also supported by the Spanish government (MIU) under grant FPU18/00516. We thank Adri\'an Javaloy and Jonas Klesen for their valuable comments and discussions. 

%% file: appendices/our_model_extended.tex
\section{Further details about \ours}
\subsection{HyperGenerator networks}
As a reminder, we work with data samples which are point clouds $(\bm{X}, \bm{Y})$ with coordinate vectors $\bm{X}$ and feature vectors $\bm{Y}$. \ours\ aims to generate a feature set $\bm{Y}$ given a set of corresponding coordinates $\bm{X}$. If the data sample is an image with $D$ pixels, $(\bm{X}, \bm{Y}$) correspond to the set of $\Rbb^2$ coordinates and values (e.g. RGB values) of the pixels, respectively.

To generate such an image, first a continuous latent variable $\zb$ is sampled from a prior distribution $p_{\psib_z}(\zb)$ parameterized by $\psib_z$. The resulting vector $\zb$ can be thought as a global summary vector of the output image and acts as the input to $K$ different \emph{hypergenerators}. Here, we refer as a hypergenerator to  both an MLP-based hypernetwork $g_{\phi_k}(\zb)$, with input  $\zb$ that outputs a set of parameters $\bm{\theta}_{k} = g_{\phi_k}(\zb)$; and, a data generator, $f_{\theta_k}$, parametrized by the output of the hypernetwork as shown in Figure \ref{fig:vamoh}. In other words, each of the MLP-based hypernetwork outputs the set of parameters $\bm{\theta}_{k} = g_{\phi_k}(\zb)$ that parameterizes the corresponding data generator $f_{\theta_k}$ which operates over each input coordinate $\bm{x}_{d} \in \bm{X}$ to generate corresponding feature values $\bm{y}_{d} \in \bm{Y}$, i.e. $\bm{y}_{d}=f_{\theta_k}(\bm{x}_{d})$ where $d \in [D]$.

The intuition for the usage of hypergenerators can be explained as follows: \ours\ uses a continous latent variable $\zb$ as an input to a hypernetwork. This hypernetwork is used for parameterizing a data generator, i.e. an INR, that can model a data sample. In the inference step, \ours\ can infer the latent variable $\zb$ using a PointConv based encoder. Therefore, \ours\ can perform conditional tasks by performing an inference step. Moreover, each of the conditional tasks requires only a single forward pass without any extra optimization step per samples. Lastly, $K$ different hypergenerators increase the expresiveness of \ours\ since they act as separate INRs for modeling different patterns in the output.

\subsection{ELBO derivation}
\label{app_sec:elbo}

Our objective is to maximize the evidence lower bound (ELBO) of a set of $D$ features $\bm{Y}$ given the corresponding set of coordinates $\bm{X}$. To do so, we introduce a family of variational distributions for the continuous and discrete latent variables, namely $\zb$ and $c_d $ where $ d \in [D]$:

\begin{equation}
     q_{\gammab}(\zb, \bm{C}|\bm{Y}, \bm{X}) 
 = q_{\gammabz}(\zb|\bm{Y}, \bm{X}) \prod_{d=1}^{D} q_{\gammabc}(c_{d}|\zb,\bm{y}_{d}, \xb_{d}).
\end{equation}

where 
\begin{equation}
    q_{\gammabz}(\zb|\bm{Y}, \bm{X}) = \normal{\zb | f_{\gammabz} (\bm{Y}, \bm{X}) } \quad \text{ and } 
    \quad q_{\gammabc}(c_{d}| \zb,\bm{y}_{d}, \xb_{d}) = \cat{c_{d}| f_{\gammabc}(\zb,\bm{y}_{d}, \xb_{d})}.
    \label{eq:factorization_appendix}
\end{equation}

Note that these two distributions are parameterized by $\gammabz$ and $\gammabc$, which we aim to learn. We refer to the set of all inference parameters as $\gammab = \{\gammabz, \gammabc\}$. 
Moving on to the generative distribution, it factorizes as

\begin{equation}
    p_{\psi, \phi}(\bm{Y}, \bm{C}, \bm{z} | \bm{X}) =  p_{\thetab}(\bm{Y} | \bm{X},  \bm{C}, \bm{z})  p_{\psi_c}(\bm{C} |  \bm{z}, \bm{X} )   p_{\psi_z}(\bm{z})
\end{equation}

where $ p_{\psi_c}(\bm{C} |  \bm{z}, \bm{X} ) $ is the prior over the categorical latent variable, $  p_{\psi_z}(\bm{z})$ is the prior over the continuous latent variable (parameterized by a Normalizing Flow, see \cref{app:flow_prior}), and  $p_{\thetab}(\bm{Y} | \bm{X},  \bm{C}, \bm{z})$ is the likelihood distribution over the set of features.  More in detail,  the likelihood factorizes over coordinates and we define each of them as a mixture:

\begin{equation}
    p_{\thetab}(\bm{Y} | \bm{X},  \bm{C}, \bm{z})  =  \prod_{d=1}^{D}  p_{\bm{\theta}}(\bm{y}_{d}|\xb_d, c_d, \zb),
    =\prod_{d=1}^{D} \prod_{k=1}^{K}  
\pi_{dk}^{\indic{c_{d}=k} } \, p_{\thetab_k}(\bm{y}_{d}|\xb_d) \quad \text{ where } \quad  \thetab_k = g_{\phi_k}(\zb).
\label{eq:generative_appendix}
\end{equation}

At this point, it is important to remark that we do not learn the parameters  $\thetab_k$ $k \in [K]$ of the  likelihood directly. Instead, we learn $k$ the parameters of $k$ different hypernetworks $g_{\phi_k}$ $k \in [K]$ that output the corresponding  likelihood parameters. Thus, the set of all the parameters of the generative models are $\phib = \{ \phib_k\}_{k=1}^K$ and $\psib = \{ \psi_z, \psi_c \}$.

With this choice of the inference and generative model, we define the ELBO as 
\begin{equation}
     \Lcal (\bm{Y}, \bm{X}; \psi, \phi, \gammab ) =  \E[ q_{\gammab}(\zb, \bm{C}|\bm{Y}, \bm{X})  ]{\log \frac{p_{\psi, \phi}(\bm{Y}, \bm{C}, \bm{z} | \bm{X})}{ q_{\gammab}(\zb, \bm{C}|\bm{Y}, \bm{X}) } } \leq \log p(\bm{Y} | \bm{X}).
\end{equation}

Using the factorization of the generative and inference model in \cref{eq:factorization_appendix,eq:generative_appendix}, we can split it in various terms:
\begin{align}
     \Lcal (\bm{Y}, \bm{X}; \psi, \phi, \gammab ) & =  \E[ q_{\gammab}(\zb, \bm{C}|\bm{Y}, \bm{X})  ]{\log  p_{\thetab}(\bm{Y} | \bm{X},  \bm{C}, \bm{z})  } \nonumber \\ 
     & +  \E[ q_{\gammab}(\zb, \bm{C}|\bm{Y}, \bm{X})  ]{\log \frac{ p_{\psi_z}(\bm{z})}{ q_{\gammab_z}(\zb |\bm{Y}, \bm{X}) } } \nonumber \\
    &  +  \E[ q_{\gammab}(\zb, \bm{C}|\bm{Y}, \bm{X})  ]{\log \frac{ p_{\psi_c}(\bm{C} |  \bm{z}, \bm{X} )}{ q_{\gammab_c}( \bm{C}|\bm{z}, \bm{Y}, \bm{X}) } }
\end{align}

We can marginalize some variables from the expectations.  
\begin{align}
     \Lcal (\bm{Y}, \bm{X}; \psi, \phi, \gammab ) & =  \E[ q_{\gammab}(\zb, \bm{C}|\bm{Y}, \bm{X})  ]{\log  p_{\thetab}(\bm{Y} | \bm{X},  \bm{C}, \bm{z})  } \nonumber\\
     & +  \E[ q_{\gammab_z}(\zb |\bm{Y}, \bm{X})  ]{\log \frac{ p_{\psi_z}(\bm{z})}{ q_{\gammab_z}(\zb |\bm{Y}, \bm{X}) } } \nonumber\\
    &  +  \E[ q_{\gammab}(\zb, \bm{C}|\bm{Y}, \bm{X})  ]{\log \frac{ p_{\psi_c}(\bm{C} |  \bm{z}, \bm{X} )}{ q_{\gammab_c}( \bm{C}|\bm{z}, \bm{Y}, \bm{X}) } }
\end{align}

And rewriting the last two terms as KL divergences we get:

\begin{align}
     \Lcal (\bm{Y}, \bm{X}; \psi, \phi, \gammab ) & =  \E[ q_{\gammab}(\zb, \bm{C}|\bm{Y}, \bm{X})  ]{\log  p_{\thetab}(\bm{Y} | \bm{X},  \bm{C}, \bm{z})  } \nonumber\\
     & - \kld{q_{\gamma_{z}}(\zb | \bm{X}, \bm{Y} )}{p_{\psi_{z}}(\zb)} \nonumber\\
 & - \E[ q_{\gammab}(\zb|\bm{Y}, \bm{X}) ] {\kld{q_{\gamma_{c}}(\bm{C} |\zb, \bm{X}, \bm{Y} )}{p_{\psi_{c}}(\bm{C} | \zb,  \bm{X})}}.
\end{align}

Finally, in the first term, we can marginalize over the categorical latent variable such that we only need to approximate by Monte Carlo expectation over $\zb$:

\begin{align}
      \E[ q_{\gammab}(\zb, \bm{C}|\bm{Y}, \bm{X})  ]{\log  p_{\thetab}(\bm{Y} | \bm{X},  \bm{C}, \bm{z})  } & =   \E[q_{\gammab}(\zb, \bm{C}|\bm{Y}, \bm{X})  ]{ \sum_{d=1}^{D} \log  p_{\thetab}(\bm{y}_{d}|\xb_d, \zb, c_d)  }   \nonumber\\
      & =  \sum_{d=1}^{D} \E[ q_{\gammab_z}(\zb |\bm{Y}, \bm{X})  ]{   \sum_{k=1}^{K} \log  p_{\thetab_k}(\bm{y}_{d}|\xb_d) \cdot \pi_{dk} } 
\end{align}

Then ELBO becomes
\begin{align}
     \Lcal (\bm{Y}, \bm{X}; \psi, \phi, \gammab ) & = \sum_{d=1}^{D} \E[ q_{\gammab_z}(\zb |\bm{Y}, \bm{X})  ]{   \sum_{k=1}^{K} \log  p_{\thetab_k}(\bm{y}_{d}|\xb_d) \cdot \pi_{dk} }  \nonumber\\
     & - \kld{q_{\gamma_{z}}(\zb | \bm{X}, \bm{Y} )}{p_{\psi_{z}}(\zb)} \nonumber\\
 & - \E[ q_{\gammab}(\zb|\bm{Y}, \bm{X}) ] {\kld{q_{\gamma_{c}}(\bm{C} |\zb, \bm{X}, \bm{Y} )}{p_{\psi_{c}}(\bm{C} | \zb,  \bm{X})}}.
\end{align}

\newpage
\subsection{Algorithm details}
The training methodology is outlined in Algorithm \ref{alg:minibatch}. As discussed in Section \ref{sec:training}, the training process begins with an initial warming stage, in which a standard prior $p(\zb)$ is utilized to stabilize the optimization of the encoder. The second and primary stage involves the introduction of the learnable prior $p_{\psi_z}(\zb)$.

\begin{algorithm}[H]
   \caption{Minibatch training of \ours}
   \label{alg:minibatch}
\begin{algorithmic}[1]
    \STATE Define hyperparameters, estimator $\widetilde{\mathcal{L}}$ using Eq. \eqref{eq:elbo_ours}
    
    \STATE $\gamma_{z,c}$ (inference), $\psi_{z,c}$ (prior), $\phi_{1:K}$ (hypernetwork) $\gets$ initialize
    \REPEAT
    \STATE $\{\bm{X}, \bm{Y}\}_{1:M} \gets$ random minibatch of $M$
    \STATE $\{\bm{X'}, \bm{Y'}\}_{1:M} \gets$
    apply dropout with probability $p\sim U(0, \alpha)$
    \IF{NF is available}
    \STATE parameterize $p_{\psi}(\zb)$ with planar flow
    \ELSE{}
    \STATE fix $p(\zb)$ with standard Gaussian
    \ENDIF
    \STATE $\mathbf{g} \gets \nabla_{\gamma, \psi, \phi_{1:K}} \widetilde{\mathcal{L}}^M\left(\gamma, \psi, \phi_{1:K}; \{\bm{X}, \bm{Y}\}_{1:M}\right)$ (gradients)
    \STATE $\gamma, \psi, \phi_{1:K}\gets$ Update parameters using gradients $\mathbf{g}$
\UNTIL convergence of parameters $\gamma, \psi, \phi_{1:K}$
\end{algorithmic}
\end{algorithm}

\subsection{Point dropout for conditional generation} \label{app:point_dropout}
\begin{figure*}[h!]
\centering
\begin{subfigure}{.47\textwidth}
    \centering
    \includegraphics[width=0.9\columnwidth]{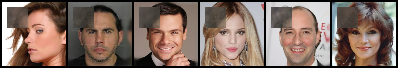}
    \caption{Without point dropout}\label{fig:point_dropout_a}
\end{subfigure}%
\begin{subfigure}{.47\textwidth}
    \centering
    \includegraphics[width=0.9\columnwidth]{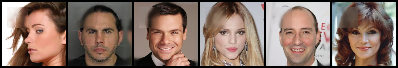}
    \caption{With point dropout}\label{fig:point_dropout_b}
\end{subfigure}%
\caption{Imputation of missing patches in test \textsc{CelebA HQ} images using VAMoH. In (a), full images are fed to the model during training. In (b), points of training images are deleted with a dropout probability per batch.}
\label{fig:point_dropout}
\end{figure*}

In this section, we present experimental evidence supporting the effectiveness of the point dropout strategy for training VAMoH, as discussed in Section \ref{sec:training} of the paper. Figure \ref{fig:point_dropout} illustrates the results of patch imputation on test images from the test set of \textsc{CelebA HQ}. When the model is trained on full images, as shown in Figure \ref{fig:point_dropout_a}, the PointConv encoder is unable to learn from a diverse set of centroids, resulting in uninformative posteriors when partial images are used during testing. However, as shown in Figure \ref{fig:point_dropout_b}, when VAMoH is trained on images that have undergone point dropout, as outlined in Section \ref{sec:training}, improved robustness for conditional generation is obtained.

\raggedbottom
\newpage
\subsection{Flow-based prior}\label{app:flow_prior}

\begin{figure*}[h!]
\centering
\begin{subfigure}{.47\textwidth}
    \centering
    \includegraphics[width=0.9\columnwidth]{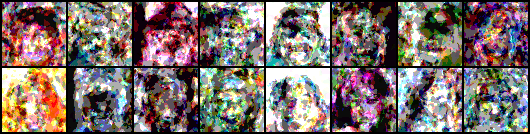}
    \caption{Standard prior, $\bm{z}\sim \mathcal{N}(\bm{0}, \bm{I})$}\label{fig:prior_samples_a}
\end{subfigure}%
\begin{subfigure}{.47\textwidth}
    \centering
    \includegraphics[width=0.9\columnwidth]{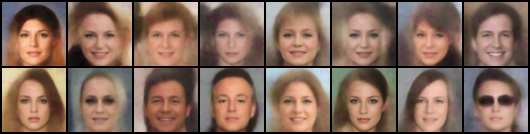}
    \caption{Standard prior, $\bm{z}\sim \mathcal{N}(\bm{0}, 0.01 \cdot\bm{I})$}\label{fig:prior_samples_b}
\end{subfigure}%
\\
\centering
\begin{subfigure}{.47\textwidth}
    \centering
    \includegraphics[width=0.9\columnwidth]{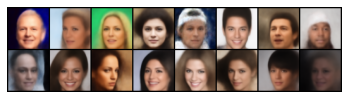}
    \caption{Standard prior,  $\bm{z}\sim q(\bm{z} | \bm{X}, \bm{Y})$}\label{fig:prior_samples_c}
\end{subfigure}%
\begin{subfigure}{.47\textwidth}
    \centering
    \includegraphics[width=0.9\columnwidth]{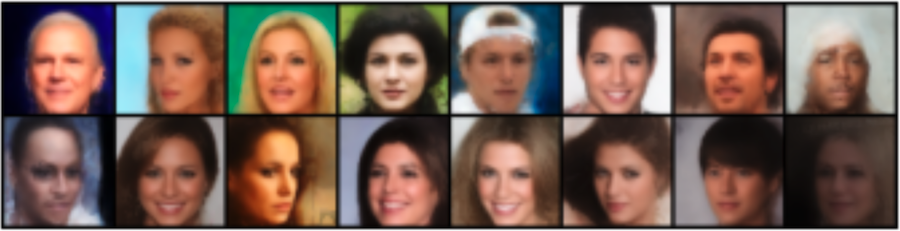}
    \caption{Flow-based prior, $\bm{z}\sim q(\bm{z} | \bm{X}, \bm{Y})$}\label{fig:prior_samples_d}
\end{subfigure}%
\\
\centering
\begin{subfigure}{.47\textwidth}
    \centering
    \includegraphics[width=0.9\columnwidth]{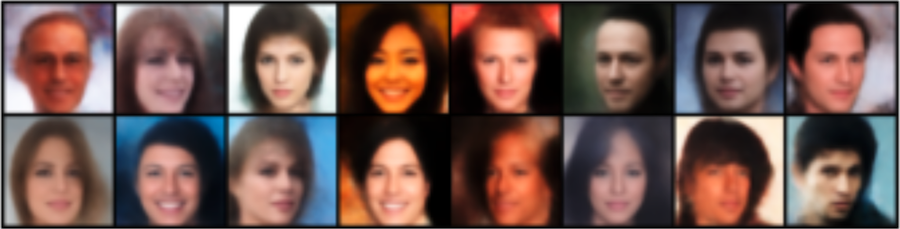}
    \caption{Flow-based prior, $\bm{z}\sim p_{\psi_z}(\bm{z})$}\label{fig:prior_samples_e}
\end{subfigure}%
\caption{Decoding samples from different latent distributions. Images in (a), (b), and (c) are generated/reconstructed by our model after being trained assuming a standard prior over $\zb$. In (d) and (e) we include reconstructions and generations, respectively, when modeling a more flexible prior over $\zb$ by training a Planar Flow.}
\label{fig:prior_samples}
\end{figure*}

In this section, we evaluate the effectiveness of the Planar Flow as a prior for our continuous latent variable in addressing the \emph{hole problem} described in Section \ref{subsec:generative_model}. We first train our model on the \celeba\ dataset \cite{liu2015faceattributes}, utilizing a fixed standard prior $p(\zb)$. Upon convergence, we sample from the learned prior and present the results in Figure \ref{fig:prior_samples_a}. The standard prior places significant probability mass in regions distant from the aggregated posterior, resulting in poor image quality when samples are drawn from these regions. By reducing the variance and approaching the posterior probability mass, as demonstrated in Figure \ref{fig:prior_samples_b}, we observe improved realism in the generated images. However, this also leads to a lack of diversity in the generated samples. In Figure \ref{fig:prior_samples_c} we include decoded samples from the posterior parameterized by the flexible encoder that reveals diversity for reconstructing images. 

In contrast, by introducing the Planar Flow after a warming stage and training its parameters, we observe a more aligned reconstruction-generation process. This is evident by comparing the samples obtained by sampling from the posterior (Figure \ref{fig:prior_samples_d}) and the learned prior (Figure \ref{fig:prior_samples_e}), respectively.

\subsection{Logistic likelihood}
Inspired by \cite{kingma2016improved}, we utilize a Discretized Logistic distribution for parameterizing the likelihood of discrete data with high number of categories, such as color channels in the RGB space for images. By doing that, we obtain a smooth and memory efficient predictive distribution for $\yb_d$, as opposed to, for example, a Categorical likelihood with parameterized with 256-way softmax. The likelihood parameterized by each hypergenerator is given by
\begin{equation}
    p_{\bm{\theta}_k}(\yb_d | \xb_d)= \sigma\left(\left(\yb_d+0.5-\bm{\mu}_{dk}\right) / \bm{s}\right)-\sigma\left(\left(\yb_d-0.5-\bm{\mu}_{dk}\right) / \bm{s}\right),
\end{equation}
 where the means $\bm{\mu}_{dk}$ are outputs of the hypergenerator, and the scales $\bm{s}$ are learnable parameters. Thanks to the design of VAMoH, our approach using a mixture for the likelihood is in line with \cite{salimans2017pixelcnn++}, and results in a Discretized Logistic Mixture Likelihood of the form
 \begin{equation}
    p(\yb_d | \xb_d)= \prod_{k=1}^K \pi_k \left[ \sigma\left(\left(\yb_d+0.5-\bm{\mu}_{dk}\right) / \bm{s}\right)-\sigma\left(\left(\yb_d-0.5-\bm{\mu}_{dk}\right) / \bm{s}\right) \right],
\end{equation}
which allows to accurately model the conditional distributions of the points by using a relatively small number of mixture components, as we show in our paper, and in concordance with \cite{salimans2017pixelcnn++}.
\subsection{Effect of mixture components}
\begin{figure}[h!]
\begin{center}
\centerline{\includegraphics[width=0.5\columnwidth]{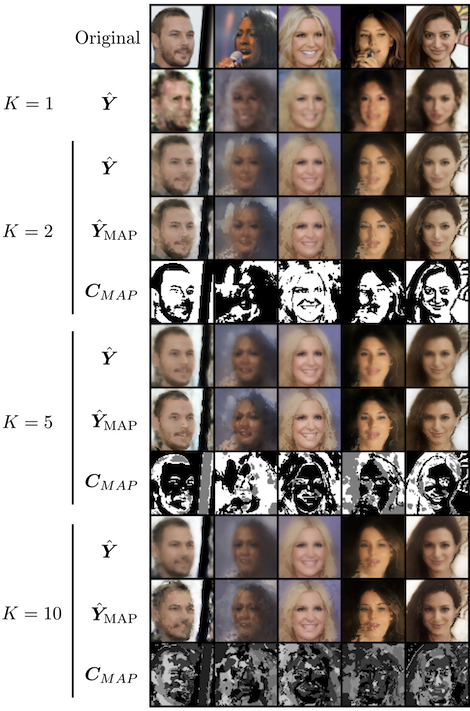}}
\caption{Ablation study on mixture of generators and effects of different number of mixture components in \ours\ for reconstruction task with \celebahq\ dataset.}
\label{fig:mixture_ablation}
\end{center}
\vskip -0.2in
\end{figure}

Using a mixture of decoders which are parameterized by hypernetworks increases the expressiveness of our model as provide a mixture of INRs to map coordinates to features in the decoding step. Furthermore, it provides the flexibility of acquiring segmentation and entropy maps as shown in Figure \ref{fig:segmentation}. In addition, in Figure
\ref{fig:mixture_ablation},  we show the effect of not using a mixture (K=1), we have lower quality outputs. Moreover, if we increase too much the number of components (K=10), we achieve similar quality; however, we get over-segmented images where regions of interest are less informative.

\raggedbottom

%% file: appendices/experiments_extended.tex
\newpage
\section{Experimental extension} \label{app:exp_ext}

In this section, we describe the experimental setup for \ours, as detailed in Appendix \ref{app:exp_setup}.
We also present a comprehensive set of results, including comparisons with the baselines \functa\ and \gasp. 
Of note, we used the original code from \gasp\ to replicate their results and also trained it on new datasets in our experimental setup. 
For \functa, we report results from the original paper \cite{dupontfuncta} for the datasets shared with our study, and for the remaining datasets, we implemented and generated results independently with our own implementation.

\subsection{Experimental setup} \label{app:exp_setup}
Implementation details for \ours\ are provided in Table \ref{table:impl_details}. We base our choice of architecture on \citep{dupont2021generative} as it provides a baseline setting. We train all of our model configurations with Adam optimizer. Also, in all of our models, we train Planar Flow and MLPs with activation of LeakyReLU with the exception of Sigmoid activation in the final layer of function generator. For datasets \celebahq, \shapesd\ and \polymnist, Discretized Logistic likelihood \cite{kingma2016improved, salimans2017pixelcnn++} is utilized, whilst for \chairs\ and \era, Bernoulli and Continuous Bernoulli \cite{loaiza2019continuous} are employed, respectively.

\begin{table*}[h!]
    \centering
\small{
\setlength\tabcolsep{3pt}
    \centering
    \caption{Implementation details of \ours.
}
    \vskip 0.1in
    \resizebox{\linewidth}{!}{
    \label{table:impl_details}
    \begin{tabular}{@{}crccccc@{}}
        \toprule
         &  & \textsc{CelebA-HQ} & \textsc{Shapes3D} & \textsc{PolyMNIST} & \textsc{ShapeNet} & \textsc{Era5} \\ \midrule
         & dim\_z & 64 & 32 & 16 & 32 & 32\\
         & K & 10 & 5 & 4 & 3 & 3 \\
         & epochs & 1000 & 600 & 600 & 500 & 600 \\
         & bs & 64 & 64 & 256 & 22 & 64 \\
         & lr & 1e-4 & 1e-4 & 1e-3 & 1e-3 & 1e-4 \\
         \midrule
        \multirow{4}{*}{\begin{tabular}[c]{@{}r@{}}PointConv\\ Encoder\end{tabular}} 
        & h\_weights & [16,16,16,16]  & [16,16] & [16,16] &[16,16,16,16]  & [16,16,16,16] \\
         & neighbors & [9,9,9,9,9] & [16,16,16] & [9,9,9] & [8,27,27,27,8] & [9,9,9,9,9] \\
         & centroids & [4096,1024,256,64,1] & [1024,256,64] & [196,49,25] & [4096,512,64,16,16] & [1024,256,64,32,16] \\
         & out\_channels & [64,128,256,512,512] & [32,64,256] & [32,32,32] & [32,64,128,256,16] & [32,64,256,512,32] \\
         & avg\_pooling\_neighbors & [9,9,9,9,None] & None & None & None &  None \\
         & avg\_pooling\_centroids & [1024,256,64,16,None] & None & None & None & None \\
          \midrule
        Categorical Encoder & layers & [64,32] & [32,32] & [32,32] & [32,32] &  [32,32]  \\
         \midrule
        Hypernetwork & layers & [256,512] & [128,256] & [16,32] & [256,512] & [256,512] \\
        \midrule
        \multirow{2}{*}{Generator} & layers & [64,64,64] & [32,32] & [4,4,4] & [64,64,64] & [64,64,64] \\
         & RFF  & $\text{m}=128, \sigma=2$ & $\text{m}=128, \sigma=2$ & $\text{m}=128, \sigma=2$ & $\text{m}=128, \sigma=2$ & $\text{m}=128, \sigma=2$ \\
         \midrule
        Flow & T & 80 & 20 & 3 & 40 & 10 \\ \bottomrule
    \end{tabular}
    }
}
\label{table:implementation_details}
\vspace{-5pt}
\end{table*}

\input{appendices/experiments_extended_generation.tex}

\input{appendices/experiments_extended_recons.tex}

\input{appendices/experiments_extended_vamoh.tex}

%% file: appendices/experiments_extended_generation.tex
\clearpage
\newpage
\subsection{Generation}
\label{app:generation}

In this section, we present a comprehensive evaluation of the generation capabilities of \ours\ in comparison to \functa\ and \gasp. \cref{tableapp:comparison} contains the obtained values for the FID, Precision, and Recall metrics for the image datasets. These values are presented as the mean and standard deviation computed over five independent sets of generated images.
For each of the metrics, we report the values using real embeddings coming from the training set (tr) and the test set (tst), since we also want to evaluate whether any model is overfitting to the training set, or as wished, it is able to generalize.

\begin{figure*}[t!]
\begin{subfigure}{.47\textwidth}
    \centering
    Original resolution
    \begin{subfigure}{0.9\columnwidth}
    \raisebox{15pt}{\rotatebox[]{90}{\small\functa}}\hspace{5pt}
        \includegraphics[width=\columnwidth]{figures/shapes3d/functa_Shapes3D_samples_1_1.0.png}
        \vspace{5pt}
        \raisebox{15pt}{\rotatebox[]{90}{\small\gasp}}\hspace{5pt}
        \includegraphics[width=\columnwidth]{figures/shapes3d/gasp_samples_shapes3d.png} 
        \vspace{5pt}
        \raisebox{12pt}{\rotatebox[]{90}{\small\ours}}\hspace{5pt}
        \includegraphics[width=\columnwidth]{figures/shapes3d/test/vamoh/vamoh_gener_x_mean_shapes3d.png}
    \end{subfigure}%
    \vspace{2pt}
    Super-resolution
    \begin{subfigure}{0.9\columnwidth}
    \raisebox{15pt}{\rotatebox[]{90}{\small\functa}}\hspace{5pt}
        \includegraphics[width=\columnwidth]{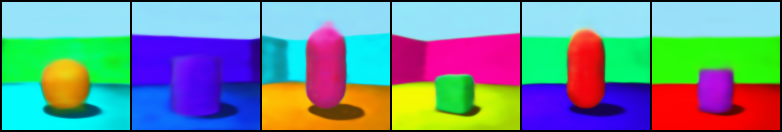}
    \raisebox{15pt}{\rotatebox[]{90}{\small\gasp}}\hspace{5pt}
        \includegraphics[width=\columnwidth]{figures/shapes3d/gasp_super_samples_shapes3d.png}
        \raisebox{12pt}{\rotatebox[]{90}{\small\ours}}\hspace{5pt}
        \includegraphics[width=\columnwidth]{figures/shapes3d/test/vamoh/vamoh_gener_super_x_mean_shapes3d.png}
    \end{subfigure}
    \caption{\shapesd}
    \label{figapp:gener_shapes}
\end{subfigure}%
\hfill
\begin{subfigure}{.47\textwidth}
    \centering
    Original resolution
    \begin{subfigure}{0.9\columnwidth}
    \raisebox{15pt}{\rotatebox[]{90}{\small\functa}}\hspace{5pt}
        \includegraphics[width=\columnwidth]{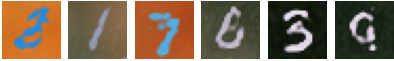}
        \vspace{5pt}
        \raisebox{15pt}{\rotatebox[]{90}{\small\gasp}}\hspace{5pt}
        \includegraphics[width=\columnwidth]{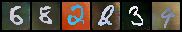}
        \vspace{5pt}
        \raisebox{12pt}{\rotatebox[]{90}{\small\ours}}\hspace{5pt}
        \includegraphics[width=\columnwidth]{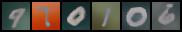}
    \end{subfigure}%
    \vspace{2pt}
    Super-resolution
    \begin{subfigure}{0.9\columnwidth}
    \raisebox{15pt}{\rotatebox[]{90}{\small\functa}}\hspace{5pt}
        \includegraphics[width=\columnwidth]{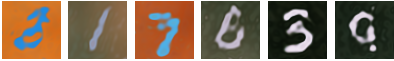}
    \raisebox{15pt}{\rotatebox[]{90}{\small\gasp}}\hspace{5pt}
        \includegraphics[width=\columnwidth]{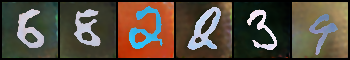}
        \raisebox{12pt}{\rotatebox[]{90}{\small\ours}}\hspace{5pt}
        \includegraphics[width=\columnwidth]{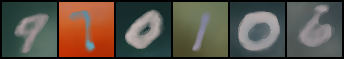}
    \end{subfigure}
    \caption{\polymnist}
    \label{figapp:gener_poly}
\end{subfigure}%
\caption{Comparison of generation quality of \ours, \functa, and \gasp\ at original resolution (top) and x2 resolution (bottom).
}
\label{figapp:gener_2}
\end{figure*}

Results for the \celebahq\ and \shapesd\ datasets were previously discussed in Section \ref{subsec:generation}. 
Additionally, we include in this Section results for the \polymnist\ dataset, where we observe that \gasp\ obtains the best quantitative results, which might be related to its ability to generate sharp samples. 
Quantitative values for \functa\ and \polymnist\ are not reported as they were not used in the original paper and we found that normalizing flows failed to avoid overfitting to the modulations, even when trying different values of dropout, number of layers, and dimensionality of the hidden space. This led to low quality of the generations, as shown in \ref{figapp:gener_poly}. 

\begin{figure*}[h!]
\begin{subfigure}{.47\textwidth}
    \centering
    Original resolution
    \begin{subfigure}{0.9\columnwidth}
    \raisebox{15pt}{\rotatebox[]{90}{\small\gasp}}\hspace{5pt}
        \includegraphics[width=\columnwidth]{figures/celeba/gasp_samples_celeba_hq.png}
        \vspace{5pt}
        \raisebox{12pt}{\rotatebox[]{90}{\small\ours}}\hspace{5pt}
        \includegraphics[width=\columnwidth]{figures/celeba/test/vamoh/vamoh_gener_x_mean_celeba_hq.png} 
    \end{subfigure}%
    \vspace{2pt}
    Super-resolution
    \begin{subfigure}{0.9\columnwidth}
    \raisebox{15pt}{\rotatebox[]{90}{\small\gasp}}\hspace{5pt}
        \includegraphics[width=\columnwidth]{figures/celeba/gasp_super_samples_celeba_hq.png} 
        \raisebox{12pt}{\rotatebox[]{90}{\small\ours}}\hspace{5pt}
        \includegraphics[width=\columnwidth]{figures/celeba/test/vamoh/vamoh_gener_super_x_mean_celeba_hq.png} 
    \end{subfigure}
    \caption{\celebahq}
    \label{figapp:gener_celebahq}
\end{subfigure}%
\hfill
\begin{subfigure}{.47\textwidth}
    \centering
    Original resolution
    \begin{subfigure}{0.9\columnwidth}
    \raisebox{15pt}{\rotatebox[]{90}{\small\gasp}}\hspace{5pt}
        \includegraphics[width=\columnwidth]{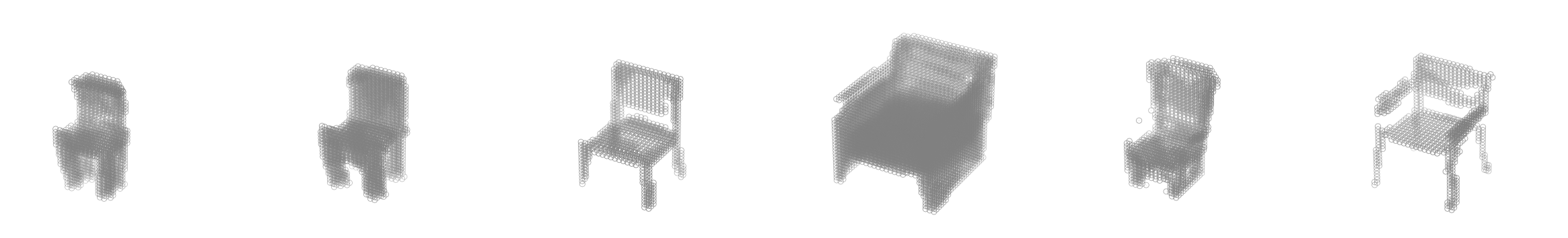}
        \vspace{5pt}
        \raisebox{12pt}{\rotatebox[]{90}{\small\ours}}\hspace{5pt}
        \includegraphics[width=\columnwidth]{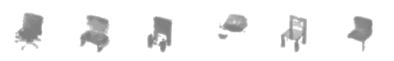} 
    \end{subfigure}%
    \vspace{2pt}
    Super-resolution
    \begin{subfigure}{0.9\columnwidth}
    \raisebox{15pt}{\rotatebox[]{90}{\small\gasp}}\hspace{5pt}
        \includegraphics[width=\columnwidth]{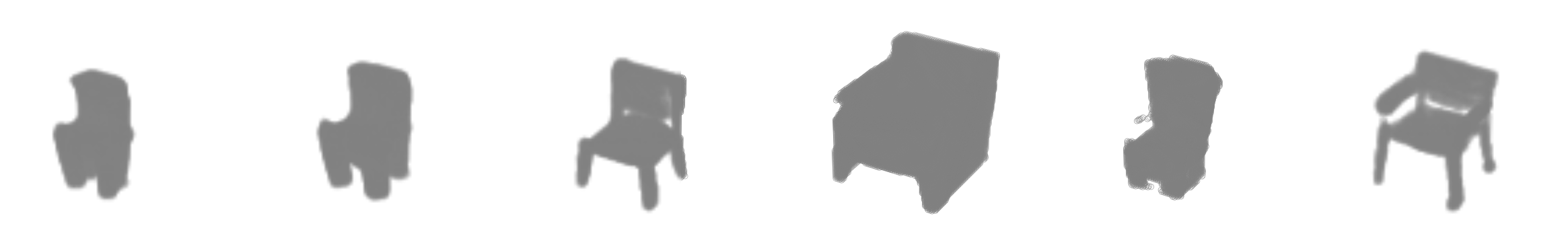}
        \raisebox{12pt}{\rotatebox[]{90}{\small\ours}}\hspace{5pt}
        \includegraphics[width=\columnwidth]{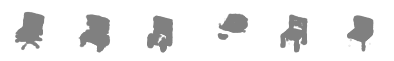} 
    \end{subfigure}
    \caption{\chairs}
    \label{figapp:gener_shapenet}
\end{subfigure}%
\caption{Comparison of generation quality of \ours\ and \gasp\ at original resolution (top) and x2 resolution (bottom).
}
\label{figapp:gener_1}
\end{figure*}

\begin{figure*}[t!]
\begin{subfigure}{.47\textwidth}
    \centering
    Original resolution
    \begin{subfigure}{0.9\columnwidth}
    \raisebox{17pt}{\rotatebox[]{90}{\small\functa}}\hspace{5pt}
        \includegraphics[width=\columnwidth]{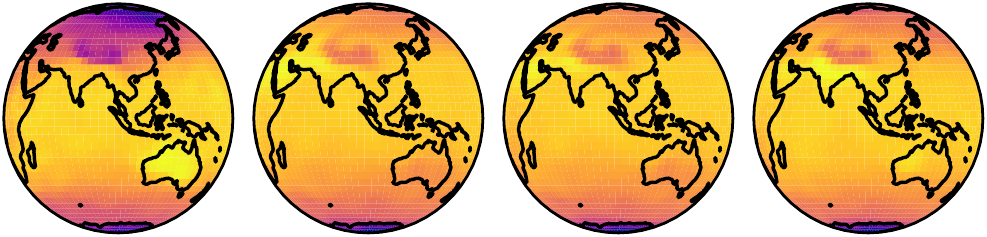}
        \vspace{5pt}
        \raisebox{17pt}{\rotatebox[]{90}{\small\gasp}}\hspace{5pt}
        \includegraphics[width=\columnwidth]{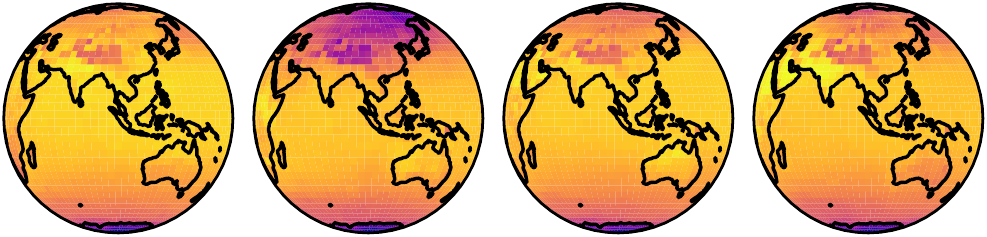}
        \vspace{5pt}
        \raisebox{15pt}{\rotatebox[]{90}{\small\ours}}\hspace{5pt}
        \includegraphics[width=\columnwidth]{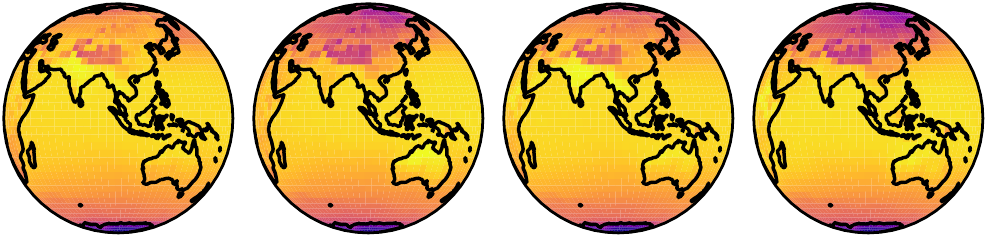}
    \end{subfigure}%
    \vspace{2pt}
\end{subfigure}%
\hfill
\begin{subfigure}{.47\textwidth}
    \centering
    Super-resolution
    \begin{subfigure}{0.9\columnwidth}
    \raisebox{17pt}{\rotatebox[]{90}{\small\functa}}\hspace{5pt}
        \includegraphics[width=\columnwidth]{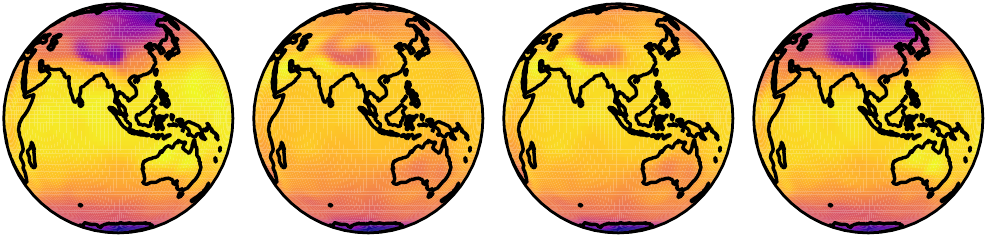}
        \vspace{5pt}
        \raisebox{17pt}{\rotatebox[]{90}{\small\gasp}}\hspace{5pt}
        \includegraphics[width=\columnwidth]{figures/era5/gasp_super_samples_era5.pdf}
        \vspace{5pt}
        \raisebox{15pt}{\rotatebox[]{90}{\small\ours}}\hspace{5pt}
        \includegraphics[width=\columnwidth]{figures/era5/vamoh_era5_test_gener_super_x_mean.pdf}
    \end{subfigure}%
    \vspace{2pt}
\end{subfigure}%
\caption{Comparison of generation quality of \ours, \functa, and \gasp\ at original resolution (left) and x2 resolution (right).
}
\label{figapp:gener_era}
\end{figure*}

Visual inspection of the generated samples in \cref{figapp:gener_1}, \cref{figapp:gener_2}  illustrates that \ours\ generates high-quality samples, capturing details such as smiles in \celebahq\  (\cref{figapp:gener_celebahq}), variety in chair legs in \chairs\ (\cref{figapp:gener_shapenet}), backgrounds in \polymnist\ (\cref{figapp:gener_poly}), as well as different configurations of temperatures (\cref{figapp:gener_era}).
In \cref{figapp:gener_1}, \cref{figapp:gener_2}, we did not include generation samples of \celebahq\ and \chairs\ for \functa\ since we were unable to reproduce the results reported in \citet{dupontfuncta}. We refer the readers to \citet{dupontfuncta} for accessing the generation results for the corresponding datasets.

\begin{table*}[h!]
    \centering
\setlength\tabcolsep{3pt}
    \centering
    \caption{
    Comparison of FID and Precision and Recall scores of image generation for \ours, \gasp, and \functa.  Low FID, high precision, and high recall indicate the best performance. The best results are highlighted in bold. Note that  for \functa\ and \celebahq\ we just report their FID value, since they do not report the precision and recall.}
    \label{tableapp:comparison}
    \vskip 0.15in
    \begin{tabular}{c ccc ccc}
\toprule
 &  \multicolumn{3}{c}{\celebahq}  &  \multicolumn{3}{c}{\shapesd}  \\
   \cmidrule(r){2-4}   \cmidrule(r){5-7}
  Model  &  $\downarrow$ FID &  $\uparrow$ Precision  &  $\uparrow$ Recall  &  $\downarrow$ FID &  $\uparrow$ Precision  &  $\uparrow$ Recall       \\
  \cmidrule(r){1-7} 
  \gasp (tst) & \bfseries 17.8 $\pm$ 0.17 & \bfseries 0.83 $\pm$ 0.0 & \bfseries 0.42 $\pm$ 0.01 & 119.35 $\pm$ 0.66 & 0.01 $\pm$ 0.0 & 0.16 $\pm$ 0.02 \\
  \functa (tst) &-  &  -  &  -  & 58.3 $\pm$ 0.14 & 0.07 $\pm$ 0.0 & 0.13 $\pm$ 0.0 \\
  \ours (tst) &72.14 $\pm$ 0.21 & 0.43 $\pm$ 0.01 & 0.0 $\pm$ 0.0 & \bfseries 56.63 $\pm$ 0.58 & 0.09 $\pm$ 0.01 & 0.63 $\pm$ 0.02 \\
  \midrule
    \gasp (tr) & \bfseries 14.01 $\pm$ 0.18 & \bfseries  0.81 $\pm$ 0.0 & \bfseries 0.43 $\pm$ 0.01 & 118.66 $\pm$ 0.64 & 0.01 $\pm$ 0.0 & 0.16 $\pm$ 0.01\\
      \functa (tr)&  40.40  &  -  &  -  & 57.81 $\pm$ 0.15 & 0.06 $\pm$ 0.0 & 0.13 $\pm$ 0.0 \\
  \ours (tr)  & 66.27 $\pm$ 0.18 & 0.65 $\pm$ 0.0 & 0.0 $\pm$ 0.0 &  \bfseries 56.25 $\pm$ 0.57 & 
 \bfseries 0.08 $\pm$ 0.0 & \bfseries 0.64 $\pm$ 0.01\\
\bottomrule
\end{tabular}
\vspace{-5pt}

\end{table*}

\raggedbottom

\raggedbottom

\raggedbottom

%% file: appendices/experiments_extended_recons.tex
\clearpage
\newpage
\subsection{Reconstruction}
\label{app:reconstruction}

\begin{figure*}[h!]
\centering
\begin{subfigure}{.5\textwidth}
    \centering
    \begin{center}
    Ground truth
    \begin{flushright}
    \includegraphics[width=0.9\columnwidth]{figures/celeba/celeba_hq_test.png}
    \end{flushright}
    Reconstructions
    \begin{subfigure}{0.9\columnwidth}
    \raisebox{15pt}{\rotatebox[]{90}{\small\functa}}\hspace{5pt}
    \includegraphics[width=\columnwidth]{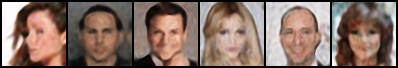}
    \raisebox{12pt}{\rotatebox[]{90}{\small\ours}}\hspace{5pt}
        \includegraphics[width=\columnwidth]{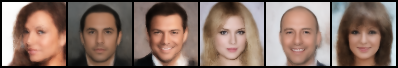}
    \end{subfigure}%
    \\
    Super-reconstructions
    \begin{subfigure}{0.9\columnwidth}
    \raisebox{15pt}{\rotatebox[]{90}{\small\functa}}\hspace{5pt}
     \includegraphics[width=\columnwidth]{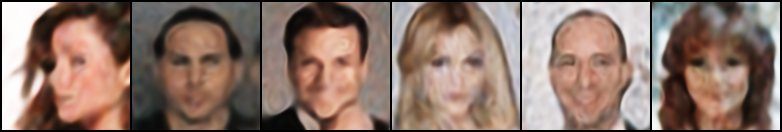}
     \raisebox{12pt}{\rotatebox[]{90}{\small\ours}}\hspace{5pt}
        \includegraphics[width=\columnwidth]{figures/celeba/test/vamoh/vamoh_recons_super_x_mean_celeba_hq.png}
    \end{subfigure}
    \end{center}
\end{subfigure}
\caption{Comparison of reconstruction quality of \ours\ and \functa\ on ground truth images of \celebahq\ dataset at original resolution (top) and x2 resolution (bottom).}
\label{figapp:recons_celeba}
\end{figure*}

\begin{figure*}[h!]
\begin{subfigure}{.47\textwidth}
    \centering
    Ground truth
    \begin{flushright}
    \includegraphics[width=0.9\columnwidth]{figures/chairs/shapenet_voxels_test_2.png}
    \end{flushright}
    Reconstructions
    \begin{subfigure}{0.9\columnwidth}
    \raisebox{15pt}{\rotatebox[]{90}{\small\functa}}\hspace{5pt}
    \includegraphics[width=\columnwidth]{figures/chairs/functa_shapenet_voxels_test_recons_0_1.0.png}
    \raisebox{12pt}{\rotatebox[]{90}{\small\ours}}\hspace{5pt}
        \includegraphics[width=\columnwidth]{figures/chairs/vamoh_shapenet_voxels_test_vamoh__voxels__recons_0.png}
    \end{subfigure}%
    \\
    Super-reconstructions
    \begin{subfigure}{0.9\columnwidth}
    \raisebox{15pt}{\rotatebox[]{90}{\small\functa}}\hspace{5pt}
     \includegraphics[width=\columnwidth]{figures/chairs/functa_shapenet_voxels_test_recons_0_2.0.png}
     \raisebox{12pt}{\rotatebox[]{90}{\small\ours}}\hspace{5pt}
        \includegraphics[width=\columnwidth]{figures/chairs/vamoh_shapenet_voxels_test_vamoh__voxels__recons_super_0.png}
    \end{subfigure}
    \caption{\chairs}
    \label{figapp:recons_right}
\end{subfigure}%
\hfill
\begin{subfigure}{.47\textwidth}
    \centering
    Ground truth
    \begin{flushright}
    \includegraphics[width=0.9\columnwidth]{figures/shapes3d/test/vamoh/recons_x_original_shapes3d.png}
    \end{flushright}
    \vspace{2pt}
    Reconstructions
    \begin{subfigure}{0.9\columnwidth}
    \raisebox{15pt}{\rotatebox[]{90}{\small\functa}}\hspace{5pt}
        \includegraphics[width=\columnwidth]{figures/shapes3d/functa_shapes3d_test_recons_0_1.0.png}
        \vspace{5pt}
        \raisebox{12pt}{\rotatebox[]{90}{\small\ours}}\hspace{5pt}
        \includegraphics[width=\columnwidth]{figures/shapes3d/test/vamoh/vamoh_recons_x_mean_shapes3d.png}
    \end{subfigure}%
    \vspace{2pt}
    Super-reconstructions
    \begin{subfigure}{0.9\columnwidth}
    \raisebox{15pt}{\rotatebox[]{90}{\small\functa}}\hspace{5pt}
        \includegraphics[width=\columnwidth]{figures/shapes3d/functa_shapes3d_test_recons_0_2.0.png}
        \raisebox{12pt}{\rotatebox[]{90}{\small\ours}}\hspace{5pt}
        \includegraphics[width=\columnwidth]{figures/shapes3d/test/vamoh/vamoh_recons_super_x_mean_shapes3d.png}
    \end{subfigure}
    \caption{\shapesd}
    \label{figapp:recons_left}
\end{subfigure}%
\\
\begin{subfigure}{.47\textwidth}
    \centering
    Ground truth
    \begin{flushright}
    \includegraphics[width=0.9\columnwidth]{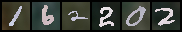}
    \end{flushright}
    \vspace{2pt}
    Reconstructions
    \begin{subfigure}{0.9\columnwidth}
    \raisebox{15pt}{\rotatebox[]{90}{\small\functa}}\hspace{5pt}
        \includegraphics[width=\columnwidth]{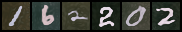}
        \vspace{5pt}
        \raisebox{12pt}{\rotatebox[]{90}{\small\ours}}\hspace{5pt}
        \includegraphics[width=\columnwidth]{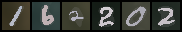}
    \end{subfigure}%
    \vspace{2pt}
    Super-reconstructions
    \begin{subfigure}{0.9\columnwidth}
    \raisebox{15pt}{\rotatebox[]{90}{\small\functa}}\hspace{5pt}
        \includegraphics[width=\columnwidth]{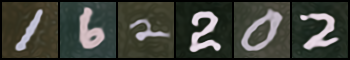}
        \raisebox{12pt}{\rotatebox[]{90}{\small\ours}}\hspace{5pt}
        \includegraphics[width=\columnwidth]{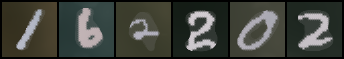}
    \end{subfigure}
    \caption{\polymnist}
    \label{fig_app:recons_poly}
\end{subfigure}%
\hfill
\begin{subfigure}{.47\textwidth}
    \centering
    Ground truth
    \begin{flushright}
    \includegraphics[width=0.85\columnwidth]{figures/era5/vamoh_era5_test_recons_x_original.pdf}
    \end{flushright}
    \vspace{2pt}
    Reconstructions
    \begin{subfigure}{0.85\columnwidth}
    \raisebox{18pt}{\rotatebox[]{90}{\small\functa}}\hspace{5pt}
        \includegraphics[width=\columnwidth]{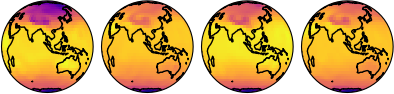}
        \vspace{5pt}
        \raisebox{15pt}{\rotatebox[]{90}{\small\ours}}\hspace{5pt}
        \includegraphics[width=\columnwidth]{figures/era5/vamoh_era5_test_recons_x_original.pdf}
    \end{subfigure}%
    \vspace{2pt}
    Super-reconstructions
    \begin{subfigure}{0.85\columnwidth}
    \raisebox{18pt}{\rotatebox[]{90}{\small\functa}}\hspace{5pt}
        \includegraphics[width=\columnwidth]{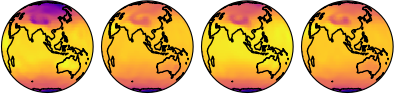}
        \raisebox{15pt}{\rotatebox[]{90}{\small\ours}}\hspace{5pt}
        \includegraphics[width=\columnwidth]{figures/era5/vamoh_era5_test_recons_super_x_mean.pdf}
    \end{subfigure}
    \caption{\era}
    \label{figapp:recons_era}
\end{subfigure}%
\caption{Comparison of reconstruction quality of \ours\ and \functa\ on ground truth data from the first samples of the test set at original and super-resolution.}
\label{figapp:recons}
\end{figure*}

In this section, we provide a comprehensive set of figures demonstrating the capability of \ours\ in reconstructing data at both the original resolution and double the original resolution. \cref{figapp:recons_celeba} and \cref{figapp:recons} present the results for all datasets under examination. These figures reveal that \ours\ is able to produce high-quality reconstructions and super-reconstructions, with visual fidelity comparable or better to that of \functa, while requiring only a simple forward pass.

Furthermore, in Section \ref{subsec:reconstruction}, the Peak Signal to Noise Ratio (PSNR) is employed to evaluate the quality of the reconstruction, denoted as $\tilde{\bm{Y}}$, of an image, denoted as $\bm{Y}$. The computation of PSNR begins by calculating the root mean squared error (RMSE) and then PSNR itself as follows
\begin{equation}
    \text{RMSE} = \sqrt{\frac{1}{D}\sum_{d} ||\yb_d - \tilde{\yb_d}||_2^2},
\end{equation}
\begin{equation}
    \text{PSNR} = 20 \log \left( \frac{255}{\text{RMSE}} \right).
\end{equation}

\begin{table*}[ht!]
    \centering
\setlength\tabcolsep{3pt}
    \centering
    \caption{Comparison of inference time (seconds)  for super-reconstruction task of \ours\ and \functa. On the right-most two columns, we show the speed improvement of VaMoH compared to Functa (3) which is trained with 3 gradient steps as suggested in the original paper \cite{dupontfuncta}
    and Functa (10) which is trained with 10 gradient steps to obtain the results of Functa depicted in Figures \ref{figapp:recons_celeba},\ref{figapp:recons}. Please note that these experiments are run on the same GPU device.}
    \label{table:time_superrecons}
    \vskip 0.15in
    \begin{tabular}{c ccc cc}
\toprule
 &  \multicolumn{3}{c}{Model Inference Time (secs)}  &  \multicolumn{2}{c}{Speed Improvement}  \\
   \cmidrule(r){2-4}   \cmidrule(r){5-6}
  Dataset  &  \ours\ &  \functa\ (3)  &  \functa\ (10)  &  vs. \functa\ (3) & vs. \functa\ (10)   \\
  \cmidrule(r){1-6} 
  \polymnist  &\bfseries0.00455 	&0.01649 	&0.05109 	&\bfseries x 3.62 	&\bfseries x 11.23
  \\
  \shapesd  &\bfseries 0.00544 	&0.01768 	&0.05489 	&\bfseries x 3.25 	&\bfseries x 10.09\\
  \celebahq  &\bfseries 0.00833 	&0.01729 	&0.05377 	&\bfseries x 2.08 	&\bfseries x 6.46\\
  \era  &\bfseries 0.00790 	&0.01997 	&0.06030 	&\bfseries x 2.53 	&\bfseries x 7.63\\
  \chairs  &\bfseries 0.01440 	&0.02089 	&0.06569 	&\bfseries x 1.45 	&\bfseries x 4.56\\
\bottomrule
\end{tabular}
\vspace{-5pt}
\end{table*}

%% file: appendices/experiments_extended_vamoh.tex
\clearpage
\newpage
\subsection{Image completion}
\label{app:im_completion}
\begin{figure*}[h!]
\begin{subfigure}{.42\textwidth}
    \centering
    \raisebox{10pt}{\rotatebox[]{90}{In}}\hspace{5pt}
    \includegraphics[width=0.91\columnwidth]{figures/image_completion/celebahq/recons_miss_x_original_occ.png}
    \vspace{2pt}
    \raisebox{12pt}{\rotatebox[]{90}{Recons.}}\hspace{5pt}
    \includegraphics[width=0.91\columnwidth]{figures/image_completion/celebahq/recons_miss_x_mean.png}
    \raisebox{10pt}{\rotatebox[]{90}{In}}\hspace{5pt}
    \includegraphics[width=0.91\columnwidth]{figures/image_completion/shapes/recons_miss_x_original_occ.png}
    \raisebox{12pt}{\rotatebox[]{90}{Recons.}}\hspace{5pt}
    \includegraphics[width=0.91\columnwidth]{figures/image_completion/shapes/recons_miss_x_mean.png}
    \raisebox{10pt}{\rotatebox[]{90}{In}}\hspace{5pt}
    \includegraphics[width=0.91\columnwidth]{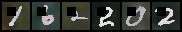}
    \raisebox{12pt}{\rotatebox[]{90}{Recons.}}\hspace{5pt}
    \includegraphics[width=0.91\columnwidth]{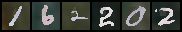}
    \raisebox{20pt}{\rotatebox[]{90}{In}}\hspace{5pt}
    \includegraphics[width=0.91\columnwidth]{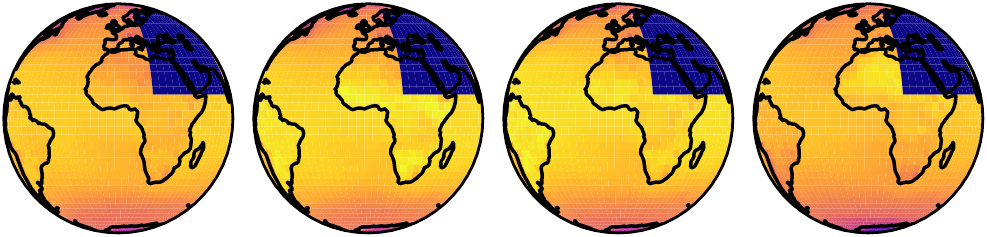}
    \raisebox{18pt}{\rotatebox[]{90}{Recons.}}\hspace{5pt}
    \includegraphics[width=0.91\columnwidth]{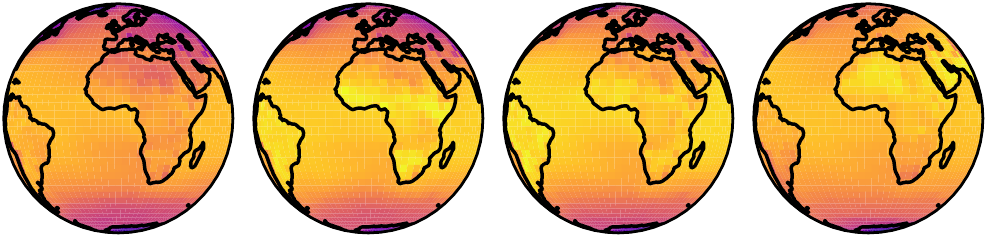}
    \caption{Missing a patch (in-painting)}
    \label{figapp:completation_left}
\end{subfigure}%
\hfill
\begin{subfigure}{.42\textwidth}
    \centering
    \raisebox{15pt}{\rotatebox[]{90}{In}}\hspace{5pt}
    \includegraphics[width=0.91\columnwidth]{figures/image_completion/celebahq/recons_half_x_original_occ.png}
    \vspace{2pt}
    \raisebox{12pt}{\rotatebox[]{90}{Recons.}}\hspace{5pt}
    \includegraphics[width=0.91\columnwidth]{figures/image_completion/celebahq/recons_half_x_mean.png} 
    \raisebox{10pt}{\rotatebox[]{90}{In}}\hspace{5pt}
    \includegraphics[width=0.91\columnwidth]{figures/image_completion/shapes/recons_half_x_original_occ.png}
    \raisebox{12pt}{\rotatebox[]{90}{Recons.}}\hspace{5pt}
    \includegraphics[width=0.91\columnwidth]{figures/image_completion/shapes/recons_half_x_mean.png}
    \raisebox{10pt}{\rotatebox[]{90}{In}}\hspace{5pt}
    \includegraphics[width=0.91\columnwidth]{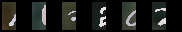}
    \raisebox{12pt}{\rotatebox[]{90}{Recons.}}\hspace{5pt}
    \includegraphics[width=0.91\columnwidth]{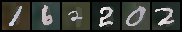}
    \raisebox{20pt}{\rotatebox[]{90}{In}}\hspace{5pt}
    \includegraphics[width=0.91\columnwidth]{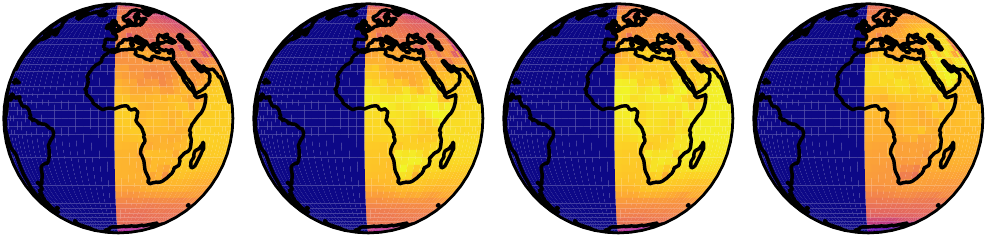}
    \raisebox{18pt}{\rotatebox[]{90}{Recons.}}\hspace{5pt}
    \includegraphics[width=0.91\columnwidth]{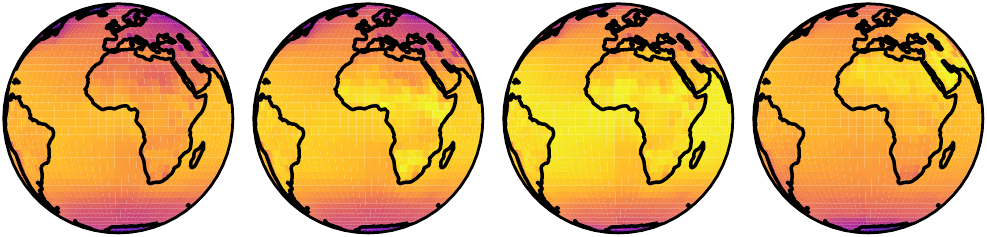}
    \caption{Missing half of the image }
    \label{figapp:completation_right}
\end{subfigure}%
\\
\begin{center}
    \begin{subfigure}{.42\textwidth}
    \centering
    \raisebox{15pt}{\rotatebox[]{90}{GT}}\hspace{5pt}
    \includegraphics[width=0.91\columnwidth]{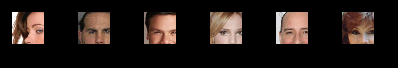}
    \vspace{2pt}
    \raisebox{12pt}{\rotatebox[]{90}{Recons.}}\hspace{5pt}
    \includegraphics[width=0.91\columnwidth]{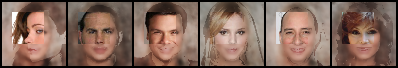}
    \raisebox{10pt}{\rotatebox[]{90}{GT}}\hspace{5pt}
    \includegraphics[width=0.91\columnwidth]{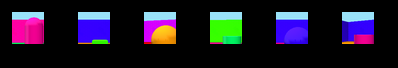}
    \raisebox{12pt}{\rotatebox[]{90}{Recons.}}\hspace{5pt}
    \includegraphics[width=0.91\columnwidth]{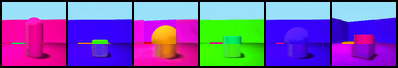}
    \raisebox{10pt}{\rotatebox[]{90}{GT}}\hspace{5pt}
    \includegraphics[width=0.91\columnwidth]{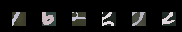}
    \raisebox{12pt}{\rotatebox[]{90}{Recons.}}\hspace{5pt}
    \includegraphics[width=0.91\columnwidth]{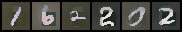}
    \caption{Image out-painting}
    \label{figapp:completation_down}
\end{subfigure}%
\end{center}
\caption{Imputation of different amounts of missing parts using \ours. For each dataset, the top row shows the input (In) and the bottom row shows the reconstructed image (Recons.). }
\label{figapp:completion}
\end{figure*}

In this section, we provide qualitative results of \ours\ for various imputation tasks for different missingness patterns. As it is already highlighted in Section \ref{sec:training} and \cref{app:point_dropout}, we use point dropout to increase the robustness of the encoder against missing data. In \cref{figapp:completion}, we provide results for image completion task on with different datasets. 

\clearpage
\newpage
\subsection{Entropy and Segmentation}
\begin{figure}[h]
\begin{subfigure}{.47\textwidth}
  \centering
  Test Recons
\includegraphics[width=\linewidth]{figures/shapes3d/test/vamoh/vamoh_recons_x_mean_shapes3d.png}
Test Entropy
\includegraphics[width=\linewidth]{figures/entropy_and_seg/shapes/recons_x_entropy.png}
Test Segmentation
\includegraphics[width=\linewidth]{figures/entropy_and_seg/shapes/recons_x_segm.png}
  \caption{\shapesd\ Test}
  \label{figapp:shapes_entropy}
\end{subfigure}%
\hfill
\begin{subfigure}{.47\textwidth}
  \centering
  Test Recons
\includegraphics[width=\linewidth]{figures/celeba/test/vamoh/vamoh_recons_x_mean_celeba_hq.png}
Test Entropy
\includegraphics[width=\linewidth]{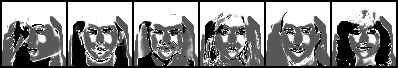}
Test Segmentation
\includegraphics[width=\linewidth]{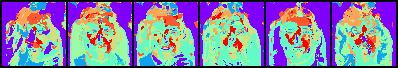}
  \caption{\celebahq\ Test}
  \label{figapp:celeba_entropy}
\end{subfigure}%
\\
\begin{center}
    \begin{subfigure}{.47\textwidth}
  \centering
  Test Recons
\includegraphics[width=\linewidth]{figures/polymnist/test/vamoh/recons_x_mean.png}
Test Entropy
\includegraphics[width=\linewidth]{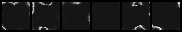}
Test Segmentation
\includegraphics[width=\linewidth]{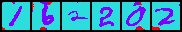}
  \caption{\polymnist\ Test}
  \label{figapp:poly_entropy}
\end{subfigure}%
\end{center}
\caption{Visualization of used components in mixture of generators of \ours\ for entropy (top) and segmentation (bottom) maps for \shapesd, \celebahq, and \polymnist.}
\label{figapp:entropy}
\end{figure}

In this section, we present figures that illustrate the contribution of each generator in the image reconstruction task. Specifically, we use entropy maps to visualize the uncertainty of the posterior probabilities of the categorical latent variable. A higher value in the entropy map indicates that the probabilities are more evenly distributed among the components for a given pixel, while a lower value suggests that a smaller number of components contribute to the reconstruction of that pixel. Additionally, we use segmentation maps to visualize the component with the highest probability per pixel, where different components are denoted by different colors. As an example, Figure \ref{figapp:celeba_entropy} shows that there is a high degree of uncertainty among the components that are responsible for reconstructing the background in the \celebahq\ dataset. Conversely, for simpler datasets such as \polymnist, as shown in Figure \ref{figapp:poly_entropy}, the components are able to differentiate between the object and background in the images. For \shapesd\ dataset results presented in Figure \ref{figapp:shapes_entropy}, the uncertainty is lower in the Regions of Interest with high variations (borders), where only one of the components focuses on generating the shape and shadow of the object, as well as the perspective of the background wall. Thus, we provide evidence that meaningful interpretations can be obtained from our Mixture of HyperGenerators.

%% file: vamoh.bbl
\begin{thebibliography}{49}
\providecommand{\natexlab}[1]{#1}
\providecommand{\url}[1]{\texttt{#1}}
\expandafter\ifx\csname urlstyle\endcsname\relax
  \providecommand{\doi}[1]{doi: #1}\else
  \providecommand{\doi}{doi: \begingroup \urlstyle{rm}\Url}\fi

\bibitem[Burgess \& Kim(2018)Burgess and Kim]{3dshapes18}
Burgess, C. and Kim, H.
\newblock 3d shapes dataset.
\newblock https://github.com/deepmind/3dshapes-dataset/, 2018.

\bibitem[Chang et~al.(2015)Chang, Funkhouser, Guibas, Hanrahan, Huang, Li,
  Savarese, Savva, Song, Su, et~al.]{chang2015shapenet}
Chang, A.~X., Funkhouser, T., Guibas, L., Hanrahan, P., Huang, Q., Li, Z.,
  Savarese, S., Savva, M., Song, S., Su, H., et~al.
\newblock Shapenet: An information-rich 3d model repository.
\newblock \emph{arXiv preprint arXiv:1512.03012}, 2015.

\bibitem[Chen et~al.(2017)Chen, Kingma, Salimans, Duan, Dhariwal, Schulman,
  Sutskever, and Abbeel]{chen2017variational}
Chen, X., Kingma, D.~P., Salimans, T., Duan, Y., Dhariwal, P., Schulman, J.,
  Sutskever, I., and Abbeel, P.
\newblock Variational lossy autoencoder.
\newblock In \emph{International Conference on Learning Representations}, 2017.
\newblock URL \url{https://openreview.net/forum?id=BysvGP5ee}.

\bibitem[Chen \& Zhang(2019)Chen and Zhang]{chen2019learning}
Chen, Z. and Zhang, H.
\newblock Learning implicit fields for generative shape modeling.
\newblock In \emph{Proceedings of the IEEE/CVF Conference on Computer Vision
  and Pattern Recognition}, pp.\  5939--5948, 2019.

\bibitem[Cremer et~al.(2018)Cremer, Li, and Duvenaud]{cremer2018inference}
Cremer, C., Li, X., and Duvenaud, D.
\newblock Inference suboptimality in variational autoencoders.
\newblock In \emph{International Conference on Machine Learning}, pp.\
  1078--1086. PMLR, 2018.

\bibitem[Dupont et~al.(2022{\natexlab{a}})Dupont, Kim, Eslami, Rezende, and
  Rosenbaum]{dupontfuncta}
Dupont, E., Kim, H., Eslami, S.~A., Rezende, D.~J., and Rosenbaum, D.
\newblock From data to functa: Your data point is a function and you can treat
  it like one.
\newblock In \emph{International Conference on Machine Learning}, pp.\
  5694--5725. PMLR, 2022{\natexlab{a}}.

\bibitem[Dupont et~al.(2022{\natexlab{b}})Dupont, Teh, and
  Doucet]{dupont2021generative}
Dupont, E., Teh, Y.~W., and Doucet, A.
\newblock Generative models as distributions of functions.
\newblock In \emph{International Conference on Artificial Intelligence and
  Statistics}, pp.\  2989--3015. PMLR, 2022{\natexlab{b}}.

\bibitem[Gatopoulos \& Tomczak(2021)Gatopoulos and Tomczak]{gatopoulos2021self}
Gatopoulos, I. and Tomczak, J.~M.
\newblock Self-supervised variational auto-encoders.
\newblock \emph{Entropy}, 23\penalty0 (6):\penalty0 747, 2021.

\bibitem[Genova et~al.(2019)Genova, Cole, Vlasic, Sarna, Freeman, and
  Funkhouser]{genova2019learning}
Genova, K., Cole, F., Vlasic, D., Sarna, A., Freeman, W.~T., and Funkhouser, T.
\newblock Learning shape templates with structured implicit functions.
\newblock In \emph{Proceedings of the IEEE/CVF International Conference on
  Computer Vision}, pp.\  7154--7164, 2019.

\bibitem[Genova et~al.(2020)Genova, Cole, Sud, Sarna, and
  Funkhouser]{genova2020local}
Genova, K., Cole, F., Sud, A., Sarna, A., and Funkhouser, T.
\newblock Local deep implicit functions for 3d shape.
\newblock In \emph{Proceedings of the IEEE/CVF Conference on Computer Vision
  and Pattern Recognition}, pp.\  4857--4866, 2020.

\bibitem[Grattarola \& Vandergheynst(2022)Grattarola and
  Vandergheynst]{grattarola2022generalised}
Grattarola, D. and Vandergheynst, P.
\newblock Generalised implicit neural representations.
\newblock \emph{Advances in Neural Information Processing Systems}, 2022.

\bibitem[Ha(2016)]{ha2016generating}
Ha, D.
\newblock Generating large images from latent vectors.
\newblock \emph{blog.otoro.net}, 2016.
\newblock URL
  \url{https://blog.otoro.net/2016/04/01/generating-large-images-from-latent-vectors/}.

\bibitem[Ha et~al.(2017)Ha, Dai, and Le]{hypernetworksha}
Ha, D., Dai, A.~M., and Le, Q.~V.
\newblock Hypernetworks.
\newblock In \emph{International Conference on Learning Representations}, 2017.
\newblock URL \url{https://openreview.net/forum?id=rkpACe1lx}.

\bibitem[Hao et~al.(2022)Hao, Mallya, Belongie, and Liu]{hao2022loe}
Hao, Z., Mallya, A., Belongie, S., and Liu, M.-Y.
\newblock Implicit neural representations with levels-of-experts.
\newblock \emph{Advances in Neural Information Processing Systems}, 2022.

\bibitem[Hersbach et~al.(2019)Hersbach, Bell, Berrisford, Biavati, Hor{\'a}nyi,
  Mu{\~n}oz~Sabater, Nicolas, Peubey, Radu, Rozum, et~al.]{era5}
Hersbach, H., Bell, B., Berrisford, P., Biavati, G., Hor{\'a}nyi, A.,
  Mu{\~n}oz~Sabater, J., Nicolas, J., Peubey, C., Radu, R., Rozum, I., et~al.
\newblock Era5 monthly averaged data on single levels from 1979 to present.
\newblock \emph{Copernicus Climate Change Service (C3S) Climate Data Store
  (CDS)}, 10:\penalty0 252--266, 2019.

\bibitem[Heusel et~al.(2017)Heusel, Ramsauer, Unterthiner, Nessler, and
  Hochreiter]{heusel2017gans}
Heusel, M., Ramsauer, H., Unterthiner, T., Nessler, B., and Hochreiter, S.
\newblock Gans trained by a two time-scale update rule converge to a local nash
  equilibrium.
\newblock \emph{Advances in neural information processing systems}, 30, 2017.

\bibitem[Hochreiter \& Schmidhuber(1997)Hochreiter and
  Schmidhuber]{hochreiter1997long}
Hochreiter, S. and Schmidhuber, J.
\newblock Long short-term memory.
\newblock \emph{Neural computation}, 9\penalty0 (8):\penalty0 1735--1780, 1997.

\bibitem[Jabbar et~al.(2020)Jabbar, Li, and Omar]{Jabbar2020ASO}
Jabbar, A., Li, X., and Omar, B.
\newblock A survey on generative adversarial networks: Variants, applications,
  and training.
\newblock \emph{ACM Computing Surveys (CSUR)}, 54, 2020.

\bibitem[Jiang et~al.(2020)Jiang, Sud, Makadia, Huang, Nie{\ss}ner, Funkhouser,
  et~al.]{jiang2020local}
Jiang, C., Sud, A., Makadia, A., Huang, J., Nie{\ss}ner, M., Funkhouser, T.,
  et~al.
\newblock Local implicit grid representations for 3d scenes.
\newblock In \emph{Proceedings of the IEEE/CVF Conference on Computer Vision
  and Pattern Recognition}, pp.\  6001--6010, 2020.

\bibitem[Karras et~al.(2017)Karras, Aila, Laine, and
  Lehtinen]{karras2017progressive}
Karras, T., Aila, T., Laine, S., and Lehtinen, J.
\newblock Progressive growing of gans for improved quality, stability, and
  variation.
\newblock \emph{arXiv preprint arXiv:1710.10196}, 2017.

\bibitem[Kingma \& Welling(2013)Kingma and Welling]{kingma2013auto}
Kingma, D.~P. and Welling, M.
\newblock Auto-encoding variational bayes.
\newblock \emph{arXiv preprint arXiv:1312.6114}, 2013.

\bibitem[Kingma et~al.(2016)Kingma, Salimans, Jozefowicz, Chen, Sutskever, and
  Welling]{kingma2016improved}
Kingma, D.~P., Salimans, T., Jozefowicz, R., Chen, X., Sutskever, I., and
  Welling, M.
\newblock Improved variational inference with inverse autoregressive flow.
\newblock \emph{Advances in neural information processing systems}, 29, 2016.

\bibitem[Klushyn et~al.(2019)Klushyn, Chen, Kurle, Cseke, and van~der
  Smagt]{klushyn2019learning}
Klushyn, A., Chen, N., Kurle, R., Cseke, B., and van~der Smagt, P.
\newblock Learning hierarchical priors in vaes.
\newblock \emph{Advances in neural information processing systems}, 32, 2019.

\bibitem[Kynk{\"a}{\"a}nniemi et~al.(2019)Kynk{\"a}{\"a}nniemi, Karras, Laine,
  Lehtinen, and Aila]{kynkaanniemi2019improved}
Kynk{\"a}{\"a}nniemi, T., Karras, T., Laine, S., Lehtinen, J., and Aila, T.
\newblock Improved precision and recall metric for assessing generative models.
\newblock \emph{Advances in Neural Information Processing Systems}, 32, 2019.

\bibitem[LeCun et~al.(1995)LeCun, Bengio, et~al.]{lecun1995convolutional}
LeCun, Y., Bengio, Y., et~al.
\newblock Convolutional networks for images, speech, and time series.
\newblock \emph{The handbook of brain theory and neural networks},
  3361\penalty0 (10):\penalty0 1995, 1995.

\bibitem[Liu et~al.(2015)Liu, Luo, Wang, and Tang]{liu2015faceattributes}
Liu, Z., Luo, P., Wang, X., and Tang, X.
\newblock Deep learning face attributes in the wild.
\newblock In \emph{Proceedings of International Conference on Computer Vision
  (ICCV)}, December 2015.

\bibitem[Loaiza-Ganem \& Cunningham(2019)Loaiza-Ganem and
  Cunningham]{loaiza2019continuous}
Loaiza-Ganem, G. and Cunningham, J.~P.
\newblock The continuous bernoulli: fixing a pervasive error in variational
  autoencoders.
\newblock \emph{Advances in Neural Information Processing Systems}, 32, 2019.

\bibitem[Ma et~al.(2020)Ma, Tschiatschek, Turner, Hern{\'a}ndez-Lobato, and
  Zhang]{ma2020vaem}
Ma, C., Tschiatschek, S., Turner, R., Hern{\'a}ndez-Lobato, J.~M., and Zhang,
  C.
\newblock Vaem: a deep generative model for heterogeneous mixed type data.
\newblock \emph{Advances in Neural Information Processing Systems},
  33:\penalty0 11237--11247, 2020.

\bibitem[Maal{\o}e et~al.(2019)Maal{\o}e, Fraccaro, Li{\'e}vin, and
  Winther]{maaloe2019biva}
Maal{\o}e, L., Fraccaro, M., Li{\'e}vin, V., and Winther, O.
\newblock Biva: A very deep hierarchy of latent variables for generative
  modeling.
\newblock \emph{Advances in neural information processing systems}, 32, 2019.

\bibitem[Mescheder et~al.(2019)Mescheder, Oechsle, Niemeyer, Nowozin, and
  Geiger]{mescheder2019occupancy}
Mescheder, L., Oechsle, M., Niemeyer, M., Nowozin, S., and Geiger, A.
\newblock Occupancy networks: Learning 3d reconstruction in function space.
\newblock In \emph{Proceedings of the IEEE/CVF conference on computer vision
  and pattern recognition}, pp.\  4460--4470, 2019.

\bibitem[Mildenhall et~al.(2021)Mildenhall, Srinivasan, Tancik, Barron,
  Ramamoorthi, and Ng]{mildenhall2021nerf}
Mildenhall, B., Srinivasan, P.~P., Tancik, M., Barron, J.~T., Ramamoorthi, R.,
  and Ng, R.
\newblock Nerf: Representing scenes as neural radiance fields for view
  synthesis.
\newblock \emph{Communications of the ACM}, 65\penalty0 (1):\penalty0 99--106,
  2021.

\bibitem[Nguyen et~al.(2021)Nguyen, Tran, Gupta, Rana, Dam, and
  Venkatesh]{nguyen2021variational}
Nguyen, P., Tran, T., Gupta, S., Rana, S., Dam, H.-C., and Venkatesh, S.
\newblock Variational hyper-encoding networks.
\newblock In \emph{Joint European Conference on Machine Learning and Knowledge
  Discovery in Databases}, pp.\  100--115. Springer, 2021.

\bibitem[Papamakarios et~al.(2021)Papamakarios, Nalisnick, Rezende, Mohamed,
  and Lakshminarayanan]{nf_prob}
Papamakarios, G., Nalisnick, E.~T., Rezende, D.~J., Mohamed, S., and
  Lakshminarayanan, B.
\newblock Normalizing flows for probabilistic modeling and inference.
\newblock \emph{J. Mach. Learn. Res.}, 22\penalty0 (57):\penalty0 1--64, 2021.

\bibitem[Peis et~al.(2022)Peis, Ma, and Hern{\'a}ndez-Lobato]{peis2022missing}
Peis, I., Ma, C., and Hern{\'a}ndez-Lobato, J.~M.
\newblock Missing data imputation and acquisition with deep hierarchical models
  and hamiltonian monte carlo.
\newblock In \emph{Advances in Neural Information Processing Systems 35}, 2022.

\bibitem[Rezende \& Mohamed(2015)Rezende and Mohamed]{Rezende2015VariationalIW}
Rezende, D. and Mohamed, S.
\newblock Variational inference with normalizing flows.
\newblock In \emph{International Conference on Machine Learning}, 2015.

\bibitem[Rezende \& Viola(2018)Rezende and Viola]{rezende2018taming}
Rezende, D.~J. and Viola, F.
\newblock Taming vaes.
\newblock \emph{arXiv preprint arXiv:1810.00597}, 2018.

\bibitem[Rezende et~al.(2014)Rezende, Mohamed, and
  Wierstra]{rezende2014stochastic}
Rezende, D.~J., Mohamed, S., and Wierstra, D.
\newblock Stochastic backpropagation and approximate inference in deep
  generative models.
\newblock In \emph{International conference on machine learning}, pp.\
  1278--1286. PMLR, 2014.

\bibitem[Rodríguez-Santana et~al.(2022)Rodríguez-Santana, Zaldivar, and
  Hernandez-Lobato]{rodriguez2022function}
Rodríguez-Santana, S., Zaldivar, B., and Hernandez-Lobato, D.
\newblock Function-space inference with sparse implicit processes.
\newblock In \emph{International Conference on Machine Learning}, pp.\
  18723--18740. PMLR, 2022.

\bibitem[Salimans et~al.(2017)Salimans, Karpathy, Chen, and
  Kingma]{salimans2017pixelcnn++}
Salimans, T., Karpathy, A., Chen, X., and Kingma, D.~P.
\newblock Pixelcnn++: Improving the pixelcnn with discretized logistic mixture
  likelihood and other modifications.
\newblock \emph{arXiv preprint arXiv:1701.05517}, 2017.

\bibitem[Simkus et~al.(2021)Simkus, Rhodes, and Gutmann]{simkus2021variational}
Simkus, V., Rhodes, B., and Gutmann, M.~U.
\newblock Variational gibbs inference for statistical model estimation from
  incomplete data.
\newblock \emph{arXiv preprint arXiv:2111.13180}, 2021.

\bibitem[Simonyan \& Zisserman(2014)Simonyan and Zisserman]{simonyan2014very}
Simonyan, K. and Zisserman, A.
\newblock Very deep convolutional networks for large-scale image recognition.
\newblock \emph{arXiv preprint arXiv:1409.1556}, 2014.

\bibitem[Sitzmann et~al.(2019)Sitzmann, Zollh{\"o}fer, and
  Wetzstein]{sitzmann2019scene}
Sitzmann, V., Zollh{\"o}fer, M., and Wetzstein, G.
\newblock Scene representation networks: Continuous 3d-structure-aware neural
  scene representations.
\newblock \emph{Advances in Neural Information Processing Systems}, 32, 2019.

\bibitem[Sitzmann et~al.(2020)Sitzmann, Martel, Bergman, Lindell, and
  Wetzstein]{sitzmann2020implicit}
Sitzmann, V., Martel, J., Bergman, A., Lindell, D., and Wetzstein, G.
\newblock Implicit neural representations with periodic activation functions.
\newblock \emph{Advances in Neural Information Processing Systems},
  33:\penalty0 7462--7473, 2020.

\bibitem[Stanley(2007)]{stanley2007compositional}
Stanley, K.~O.
\newblock Compositional pattern producing networks: A novel abstraction of
  development.
\newblock \emph{Genetic programming and evolvable machines}, 8\penalty0
  (2):\penalty0 131--162, 2007.

\bibitem[Tancik et~al.(2020)Tancik, Srinivasan, Mildenhall, Fridovich-Keil,
  Raghavan, Singhal, Ramamoorthi, Barron, and Ng]{Tancik20}
Tancik, M., Srinivasan, P., Mildenhall, B., Fridovich-Keil, S., Raghavan, N.,
  Singhal, U., Ramamoorthi, R., Barron, J., and Ng, R.
\newblock Fourier features let networks learn high frequency functions in low
  dimensional domains.
\newblock In Larochelle, H., Ranzato, M., Hadsell, R., Balcan, M., and Lin, H.
  (eds.), \emph{Advances in Neural Information Processing Systems}, volume~33,
  pp.\  7537--7547. Curran Associates, Inc., 2020.
\newblock URL
  \url{https://proceedings.neurips.cc/paper/2020/file/55053683268957697aa39fba6f231c68-Paper.pdf}.

\bibitem[Tomczak \& Welling(2018)Tomczak and Welling]{tomczak2018vae}
Tomczak, J. and Welling, M.
\newblock Vae with a vampprior.
\newblock In \emph{International Conference on Artificial Intelligence and
  Statistics}, pp.\  1214--1223. PMLR, 2018.

\bibitem[Wu et~al.(2019)Wu, Qi, and Fuxin]{wu2019pointconv}
Wu, W., Qi, Z., and Fuxin, L.
\newblock Pointconv: Deep convolutional networks on 3d point clouds.
\newblock In \emph{Proceedings of the IEEE/CVF Conference on Computer Vision
  and Pattern Recognition}, pp.\  9621--9630, 2019.

\bibitem[Zeng et~al.(2022)Zeng, Vahdat, Williams, Gojcic, Litany, Fidler, and
  Kreis]{zeng2022lion}
Zeng, X., Vahdat, A., Williams, F., Gojcic, Z., Litany, O., Fidler, S., and
  Kreis, K.
\newblock Lion: Latent point diffusion models for 3d shape generation.
\newblock In \emph{Advances in Neural Information Processing Systems
  (NeurIPS)}, 2022.

\bibitem[Zhang et~al.(2018)Zhang, B{\"u}tepage, Kjellstr{\"o}m, and
  Mandt]{zhang2018advances}
Zhang, C., B{\"u}tepage, J., Kjellstr{\"o}m, H., and Mandt, S.
\newblock Advances in variational inference.
\newblock \emph{IEEE transactions on pattern analysis and machine
  intelligence}, 41\penalty0 (8):\penalty0 2008--2026, 2018.

\end{thebibliography}
